%% file: main.tex
\documentclass[11pt]{article}
\pdfoutput=1

\usepackage{url}
\usepackage{color}
\usepackage[letterpaper, margin=3cm]{geometry}
\usepackage{todonotes}
\usepackage[frozencache]{minted}
\usepackage{amsmath}
\usepackage{etoolbox}
\usepackage{subfigure}
\usepackage{graphicx}
\usepackage{xr}
\usepackage[perpage]{footmisc}

\makeatletter
\newcommand*{\addFileDependency}[1]{
  \typeout{(#1)}
  \IfFileExists{#1}{}{\typeout{No file #1.}}
}
\makeatother

\newcommand{\ignore}[1]{}

\newcommand{\footremember}[2]{%
    \footnote{#2}
    \newcounter{#1}
    \setcounter{#1}{\value{footnote}}%
}
\newcommand{\footrecall}[1]{%
    \footnotemark[\value{#1}]%
} 

\newcommand{\secref}[1]{\S\ref{sec:#1}}
\newcommand{\tabref}[1]{Table~\ref{tab:#1}}
\newcommand{\figref}[1]{Figure~\ref{fig:#1}}

\title{DyNet: The Dynamic Neural Network Toolkit}
\author{
  Graham Neubig\footremember{cmu}{Carnegie Mellon University, Pittsburgh, PA, USA}\footremember{naist}{Nara Institute of Science and Technology, Ikoma, Japan}, \hspace{1mm}
  Chris Dyer\footremember{deepmind}{DeepMind, London, UK}, \hspace{1mm}
  Yoav Goldberg\footremember{biu}{Bar Ilan University, Ramat Gan, Israel}, \hspace{1mm}
  Austin Matthews\footrecall{cmu}, \\
  Waleed Ammar\footremember{ai2}{Allen Institute for Artificial Intelligence, Seattle, WA, USA}, \hspace{1mm}
  Antonios Anastasopoulos\footremember{nd}{University of Notre Dame, Notre Dame, IN, USA}, \hspace{1mm}
  Miguel Ballesteros\footremember{ibmtjw}{IBM T.J. Watson Research Center, Yorktown Heights, NY, USA}, \hspace{1mm}
  David Chiang\footrecall{nd}, \\
  Daniel Clothiaux\footrecall{cmu}, \hspace{1mm}
  Trevor Cohn\footremember{melbourne}{University of Melbourne, Melbourne, Australia}, \hspace{1mm}
  Kevin Duh\footremember{jhu}{Johns Hopkins University, Baltimore, MD, USA}, \hspace{1mm}
  Manaal Faruqui\footremember{googleny}{Google, New York, NY, USA}, \\
  Cynthia Gan\footremember{googlemtv}{Google, Mountain View, CA, USA}, \hspace{1mm}
  Dan Garrette\footrecall{googleny}, \hspace{1mm}
  Yangfeng Ji\footremember{uw}{University of Washington, Seattle, USA}, \hspace{1mm}
  Lingpeng Kong\footrecall{cmu}, \hspace{1mm}
  Adhiguna Kuncoro\footrecall{cmu}, \hspace{1mm} \\
  Gaurav Kumar\footrecall{jhu}, \hspace{1mm} 
  Chaitanya Malaviya\footrecall{cmu}, \hspace{1mm}
  Paul Michel\footrecall{cmu}, \hspace{1mm}
  Yusuke Oda\footrecall{naist}, \hspace{1mm} \\
  Matthew Richardson\footremember{msr}{Microsoft Research, Redmond, WA, USA}, \hspace{1mm}
  Naomi Saphra\footremember{edinburgh}{University of Edinburgh, Edinburgh, UK},
  Swabha Swayamdipta\footrecall{cmu}, \hspace{1mm}
  Pengcheng Yin\footrecall{cmu} \hspace{1mm}
}

\date{}

\begin{document}
\maketitle

\begin{abstract}
We describe DyNet, a toolkit for implementing neural network models based on dynamic declaration of network structure.
In the static declaration strategy that is used in toolkits like Theano, CNTK, and TensorFlow, the user first defines a computation graph (a symbolic representation of the computation), and then examples are fed into an engine that executes this computation and computes its derivatives.
In DyNet's dynamic declaration strategy,  computation graph construction is mostly transparent, being implicitly constructed by executing procedural code that computes the network outputs, and the user is free to use different network structures for each input.
Dynamic declaration thus facilitates the implementation of more complicated network architectures, and DyNet is specifically designed to allow users to implement their models in a way that is idiomatic in their preferred programming language (C++ or Python).
One challenge with dynamic declaration is that because the symbolic computation graph is defined anew for every training example, its construction must have low overhead. 
To achieve this, DyNet has an optimized C++ backend and lightweight graph representation. Experiments show that DyNet's speeds are faster than or comparable with static declaration toolkits, and significantly faster than Chainer, another dynamic declaration toolkit.
DyNet is released open-source under the Apache 2.0 license, and available at \url{http://github.com/clab/dynet}.
\end{abstract}

\section{Introduction}
\label{sec:intro}
\input{010-intro}
\addFileDependency{010-intro}

\section{Static Declaration vs. Dynamic Declaration}
\label{sec:staticvsdynamic}
\input{015-staticvsdynamic}
\addFileDependency{015-staticvsdynamic}

\section{Coding Paradigm}
\label{sec:codingparadigm}
\input{020-codingparadigm}
\addFileDependency{020-codingparadigm}

\section{Behind the Scenes}
\label{sec:design}
\input{030-design}
\addFileDependency{030-design}

\section{Higher Level Abstractions}
\label{sec:abstractions}
\input{040-abstractions}
\addFileDependency{040-abstractions}

\section{Efficiency Tools}
\label{sec:efficiency}
\input{045-efficiency}

\addFileDependency{045-efficiency}

\section{Empirical Comparison}
\label{sec:eval}
\input{050-eval}
\addFileDependency{050-eval}

\section{Use Cases}
\label{sec:usecases}
\input{055-usecases}
\addFileDependency{055-usecases}

\section{Conclusion}
\label{sec:conclusion}
\input{060-conclusion}
\addFileDependency{060-conclusion}

\section{Acknowledgements}
\label{sec:acks}
\input{070-acks}
\addFileDependency{070-acks}

{\small
\bibliographystyle{plain}
\bibliography{confnames,dynet-paper}
}

\end{document}

%% file: 010-intro.tex
Deep neural networks are now an indispensable tool in the machine learning practitioner's toolbox, powering applications from image understanding \cite{krizhevsky2012imagenet}, speech recognition and synthesis \cite{hinton2012deep,zen2013statistical}, game playing \cite{mnih2015human,silver2016mastering}, language modeling and analysis \cite{bengio2003neural,collobert2011natural}, and more. To a first approximation, deep learning replaces the application-specific feature engineering (coupled with well-understood models) that is characteristic of classical ``shallow'' learning paradigms with application-specific model engineering (usually coupled with less sophisticated features of the inputs). The deep learning paradigm therefore entails ongoing development of new model variants. While developing effective models requires insight and analysis, it also requires implementing new models and assessing their performance on real tasks. Thus, rapid prototyping and easy maintenance of efficient and correct model code is of paramount importance in deep learning.

Deep learning models operate in two modes: they either compute a prediction (or distribution over predictions) given an input, or, at training time when supervision is available, they compute the derivatives of a prediction error ``loss'' with respect to model parameters which are used to minimize subsequent errors on similar inputs using some variant of gradient descent. Since implementing a model requires both implementing the code that executes the model predictions as well as the code that carries out gradient computation and learning, model development is a nontrivial engineering challenge. The difficulty of this challenge can be reduced by using tools that simplify implementation of neural network computations. These tools, including Theano \cite{bergstra2010theano}, TensorFlow \cite{abadi2016tensorflow}, Torch \cite{collobert2002torch}, CNTK \cite{yu2014cntk}, MxNet \cite{chen2015mxnet}, and Chainer \cite{tokui2015chainer} provide neural network function primitives (e.g., linear algebra operations, nonlinear transformations, etc.), parameter initialization and optimization routines, and usually the ability to express composite computations of a task-specific prediction and error that are then differentiated automatically to obtain the gradients needed to drive learning algorithms. This last automatic differentiation (autodiff) component is arguably their most important labor-saving feature since changes to the function that computes the loss for a training input will require a corresponding change in the computation of its derivative. If an engineer maintains these code paths independently, it is easy for them to get out of sync. Furthermore, since the differentiation of composite expressions is algorithmically straightforward~\cite{wengert1964asa,hogan2014fast}, there is little downside to using autodiff algorithms in place of hand-written code for computing derivatives.

In short, these tools have made deep learning successful because they have effectively solved a crucial software engineering problem. The question remains though: since engineering is a key component of deep learning practice, what engineering problems are existing tools failing to solve? Do they let programmers express themselves naturally? Do they facilitate debugging? Do they facilitate the maintenance of large-scale projects?

In this paper, we suggest that the programming model that underlies several popular toolkits---namely a separation of declaration and execution of the network architecture (which we refer to as \emph{static declaration})---is necessarily associated with some serious software engineering risks, particularly when dealing with dynamically structured network architectures (e.g., sequences of variable lengths and tree-structured recursive neural networks). As an alternative, we propose as reviving an alternative programming model found in autodiff libraries that unifies declaration and execution.

As a proof of concept of our recommended programming model, we describe ``DyNet: The Dynamic Neural Network Toolkit,'' a toolkit based on a unified declaration and execution programming model which we call \textit{dynamic declaration}.\footnote{Available open-source under the Apache license at \url{https://github.com/clab/dynet} and documented at \url{https://dynet.readthedocs.io}.}

In a series of case-studies in a single-machine environment,\footnote{Static declaration facilitates algorithmic distribution of computation across multiple devices, but our focus is in this paper is on the engineering challenges associated with rapid model prototyping. We discuss static declaration's benefits with respect to scalability below, and we will return to the outlook for automatic distribution of dynamic computation across multiple devices in the conclusion.} we show that DyNet obtains execution efficiency that is comparable to static declaration toolkits for standard model architectures. For models that use dynamic architectures (e.g., where every training instance has a different model architecture), implementation is significantly simplified.

%% file: 015-staticvsdynamic.tex
In this section, we describe in more concrete terms the two paradigms of static declaration~(\S\ref{sec:staticdecl}) and dynamic declaration~(\S\ref{sec:dynamicdecl}) .

\subsection{Static Declaration}
\label{sec:staticdecl}

Programs written in the static declaration paradigm (which is used by TensorFlow, Theano, and CNTK, among others) follow a two step process:
\begin{description}
\item[Definition of a computational architecture:] The user writes a program that defines the ``shape'' of the computation they wish to perform. This computation is generally represented as a \emph{computation graph}, which is a symbolic representation of a complex computation that can both be executed and differentiated using autodiff algorithms. For example, a user might define computation that performs image recognition such as: ``Take a 64x64 image represented as a matrix of brightness values, perform two layers of convolutions, predict the class of the image out of 100 classes, and optionally calculate the loss with respect to the correct label.''
\item[Execution of the computation:] The user then repeatedly populates the 64x64 matrix with data and the library executes the previously declared computation graph. The predictions can then be used at test time, or at training time the loss can be calculated and back-propagated through the graph to compute the gradients required for parameter updates.
\end{description}
Static declaration has a number of advantages, the most significant of which is that after the computation graph is defined, it can be optimized in a number of ways so that the subsequent repeated executions of computation can be performed as quickly as possible, speeding training and test over large datasets. Second, the static computation graph can be used to schedule computation across a pool of computational devices~\cite{abadi2016tensorflow}. Third, the static declaration paradigm benefits the toolkit designer: less efficient algorithms for graph construction and optimization can be used since this one-time cost will be amortized across many training instances at run time.
These advantages are the reason why many popular libraries take this static declaration approach.

Unfortunately, there are a couple of reasons why this variety of static declaration can also be inconvenient, specifically in situations where we would like to use networks in which each piece of data requires a different architecture.
Here are a few examples:
\begin{description}
\item[Variably sized inputs:]
Analogously to the previous example, if we are not restricted to 64x64 images, but have a different sized input for each image, it is more difficult to define a single structure of identical computations.
We also can think of a situation where we want to process a natural language sentence to perform machine translation \cite{sutskever14sequencetosequence,bahdanau15alignandtranslate}, where each sentence is of a different size and we need to run a recurrent neural network (RNN) \cite{elman90rnn,hochreiter97lstm} over a variable number of words to process it correctly. 
\item[Variably structured inputs:]
A more complicated case is when each input has not only a different size, but also a different structure, such as the case of tree-structured networks~\cite{socher11recursivenn,tai15treelstm} or graph-structured networks~\cite{liang:2016eccv}, which have been used in a variety of tasks.
In these networks, each input data point may have a different structure, which means that computation will be different for different examples, substantially complicating the use of a fixed architecture and computation order.
\item[Nontrivial inference algorithms:]
In some cases, it is desirable to use sophisticated inference algorithms to compute or approximate a differentiable quantity such as the marginal likelihood or Bayes risk during learning~\cite{graves:2006,do:2010,gormley:2015,kong2016segrnn}. These algorithms may require structured intermediate values (e.g., dynamic programming charts or graphs in belief propagation) and may use nontrivial flow control.
\item[Variably structured outputs:]
Finally, the structure of the output could change, even based on the values calculated by earlier steps in the network.
This is true for applications such as syntactic parsing using tree-structured networks \cite{dyer2015stacklstm}, or predictive models that use variably sized search spaces based on the certainty of the model at any particular decision point \cite{buckman2016backtracking}.
Handling these outputs necessitates the ability to perform dynamic flow control.
\end{description}

The above-mentioned cases are difficult for simple static declaration tools to handle. However, by increasing the power and complexity of the computational graph formalism, it is possible to support them.
For example, it is possible to process variable sized inputs if the computation graph can represent objects whose size is unspecified at declaration time. To cope with structured data, flow control operations such as conditional execution, iteration, etc., can be added to the inventory of operations supported by the computation graph. For example, to run an RNN over variable length sequences, Theano offers the \texttt{scan} operation, and TensorFlow offers the \texttt{dynamic\_rnn} operation.

While it is therefore possible to deal with variable architectures with static declaration in principle, it still poses some difficulties in practice:
\begin{description}
\item[Difficulty in expressing complex flow-control logic:] When computing the value of a loss function requires traversal of complex data structures or the execution of nontrivial inference algorithms, it is necessary, in the static declaration paradigm, to express the logic that governs these algorithms as part of the computation graph. However, in existing toolkits, iteration, recursion, and even conditional execution (standard flow control primitives in any high-level language) look very different in the graph than in the imperative coding style of the host language---if they are present at all~\cite{looks2017dynamic,bowman2016spinn}. Since traversal and inference algorithms can be nontrivial to implement correctly under the best of circumstances~\cite{eisner:2005}, implementing them ``indirectly'' in a computation graph requires considerable sophistication on the developer's part. 
\item[Complexity of the computation graph implementation:] To support dynamic execution, the computation graph must be able to handle more complex data types (e.g., variable sized tensors and structured data), and operations like flow control primitives must be available as operations. This increases the complexity of computation graph formalism and implementation, and reduces opportunities for optimization.
\item[Difficulty in debugging:] While static analysis permits some errors to be identified during declaration, many logic errors will necessarily wait to be uncovered until execution (especially when many variables are left underspecified at declaration time), which is necessarily far removed from the declaration code that gave rise to them. This separation of the location of the root cause and location of the observed crash makes for difficult debugging.
\end{description}
These difficulties highlight the need for an alternative paradigm for implementing neural networks that is simple and intuitive to program and debug, in a language widely known and used by programmers.

\subsection{Dynamic Declaration}
\label{sec:dynamicdecl}

In contrast to the two-step process of definition and execution used by the static declaration paradigm, the dynamic declaration model that we advocate in this paper takes a single-step approach, in which the user defines the computation graph programmatically as if they were calculating the outputs of their network on a particular training instance.
For example, in the case of image processing from above, this would mean that for every training example, the user would ``Load a 64x64 image into their computation graph, perform several convolutions, and calculate either the predictive probabilities (for test) or loss (for training).''
Notably, there are no separate steps for definition and execution: the necessary computation graph is created, on the fly, as the loss calculation is executed, and a new graph is created for each training instance. This requires very lightweight graph construction.

This general design philosophy of implicit graph construction has been around in the form of autodiff libraries since at least the 1980's \cite{hogan2014fast,griewank1991automatic} and, in addition to DyNet, is implemented in the neural network library Chainer \cite{tokui2015chainer}.
It is advantageous because it allows the user to: (1)~define a different computation architecture for each training example or batch, allowing for the handling of variably sized or structured inputs using flow-control facilities of the host language, and (2)~interleave definition and execution of computation, allowing for the handling of cases where the structure of computation may change depending on the results of previous computation steps. Furthermore, it reduces the complexity of the computation graph implementation since it does not need to contain flow control operations or support dynamically sized data---these are handled by the host language (C++ or Python).
In \secref{codingparadigm} we discuss the coding paradigm in more depth, and give examples of how these cases that are difficult when using static computation are handled simply within DyNet.

Despite the attractiveness of the dynamic declaration paradigm, there is a reason why many of the major neural network libraries opt for static declaration: creating and optimizing computation graphs can be expensive, and by spreading this cost across many training instances, the amortized cost of even an inefficient implementation will be negligible.
The over-arching goal of DyNet is to close this gap by minimizing the computational cost of graph construction, allowing for efficient dynamic computation, and removing barriers to rapid prototyping and implementation of more sophisticated applications of neural nets that are not easy to implement in the static computation paradigm.
In order to do so, DyNet's backend, which is written in C++, is optimized in a number of ways to remove overhead in computation graph construction, and support efficient execution on both CPU and GPU. This optimization is not as tricky as might at first seem: because flow control and facilities for dealing with variably sized inputs remain in the host language (rather than in the computation graph, as is required by static declaration), the computation graph needs to support fewer operation types, and these tend to be more completely specified (e.g., tensor sizes are always known rather than inferred at execution time). This results in optimization opportunities and keeps the library code simple. The design of the backend is explained more comprehensively in \secref{design}.

DyNet is also designed to have native support for a number of more complicated use cases that particularly come up in natural language processing or other sequential tasks.
For example, it is common to use recurrent neural networks \cite{elman90rnn,hochreiter97lstm}, or tree-structured neural networks \cite{socher11recursivenn,tai15treelstm}, both of which are supported natively by DyNet (\secref{abstractions}, c.f. some other toolkits such as Theano, which often require a wrapper using a separate tool such as Keras\footnote{\url{https://keras.io}}).
There is also back-end support for mini-batching (\secref{minibatching}) to improve computational efficiency, taking much of the mental burden off of users who want to implement mini-batching in their models.
Finally, for more complicated models that do not support mini-batching, there is also support for data-parallel multi-processing (\secref{parallelization}), in which asynchronous parameter updates are performed across multiple threads, making it simple to parallelize (on a single machine) any variety of model at training time.

%% file: 020-codingparadigm.tex
\subsection{Coding Paradigm Overview}
\label{sec:paradigmoverview}

From the user's perspective, writing a program using DyNet revolves around
building up \texttt{Expression}s that correspond to the computation that needs to
be performed. This first starts with basic expressions that are either
constant input values, or model \texttt{Parameters}. Then, compound expressions
are further built from other expressions by means of
\texttt{operations}, and the chain of operations implicitly define a
\texttt{ComputationGraph} for the desired computation.
This computation graph represents symbolic computation, and the results of the
computation are evaluated lazily: the
computation is only performed once the user explicitly asks for it (at which
point a ``forward'' computation is triggered). Expressions that evaluate to
scalars (i.e. loss values) can also be used to trigger a ``backward'' computation,
computing the gradients of the computation with respect to the parameters.
The parameters and gradients are saved in a \texttt{Model} object, and a
\texttt{Trainer} is used to update the parameters according to the gradients and
an update rule.

We briefly elaborate on each of these components below:
\begin{description}

\item[Parameter and LookupParameter:] \texttt{Parameter}s are vectors, matrices
  or tensors of real numbers representing things like weight matrices and bias
  vectors. \texttt{LookupParameter}s are sets of vectors of parameters that we
  intend to look up sparsely, such as word embeddings. I.e., if we have a vocabulary
  $V$ for which we would like to look up embeddings,
  a \texttt{LookupParameters} object defines a $|V|\times d$ matrix, which acts as
  an embedding matrix mapping items in $0,\ldots ,|V|-1$ to $d$-dimensional vectors.
  Parameters and LookupParameters are stored in a Model, and persist across
  training examples (i.e., across different ComputationGraph instances).
  
\item[Model:] A model is a collection of \texttt{Parameter}s and
  \texttt{LookupParameter}s. The user obtains the Parameters by requesting them
  from a model. The model then keeps track of the parameters (and their
  gradients). Models can be saved to and loaded from disk, and are also used by
  the Trainer object below.

\item[Trainer:] A trainer implements an online update rule such as simple stochastic gradient descent,
  AdaGrad \cite{duchi2011adaptive}, or Adam \cite{kingma2014adam}.
  The \texttt{Trainer} holds a pointer to the Model object, and hence the
  parameters within it, and also may maintain other information about the
  parameters as required by the update rule.

\item[Expression:] Expressions are the main data types being manipulated in a DyNet
  program. Each expression represents a sub-computation in a computation graph.
  For example, a \texttt{Parameter}s object representing a matrix or a vector can be added
  to the ComputationGraph resulting in an expression $W$ or $b$. Similarly, a
  \texttt{LookupParameters} object $E$ can be queried for a specific embedding
  vector (which is added to the computation graph) by a lookup operation,
  resulting in an Expression $E[i]$. These Expressions can then be combined into
  larger expressions, such as $\text{concatenate}(E[3],E[4])$ or $\text{softmax}(\tanh(W * \text{concatenate}(E[3],E[4]) + b))$.
  Here, $\text{softmax}$, $\tanh$, $*$, $+$, $\text{concatenate}$ are
  operations, discussed below.

\item[Operations:] These are not objects, but rather functions that operate on
  expressions and return expressions, building a computation graph in the
  background.
  DyNet define operations for many basic
  arithmetic primitives (addition, multiplication, dot-product, softmax, \ldots) as
  well as common loss functions, activation functions, and so on. When
  applicable, the operations are defined using operator overloading, making
  graph construction as natural and intuitive as possible.

\item[Builder Classes:] Builder classes define interfaces for creating various
  ``standardized'' network components, such as recurrent neural network,
  tree-structured network, and large-vocabulary softmax. These work on top of
  expressions and operations, and provide easy-to-use libraries. Builder
  classes provide convenient and efficient implementations of standard
  algorithms, but are not part of the ``core'' DyNet library, in the sense that the builders
  are higher-level constructs that are implemented ``on top of'' the core DyNet
  auto-differentiation functionality.
  Builders are discussed more in depth in \secref{abstractions} below.
  
\item[ComputationGraph:] Expressions are part of an implicit \texttt{ComputationGraph}
  object that
  defines the computation that needs to be performed. DyNet currently assumes that
  only one computation graph will exist at any one time.
  While the ComputationGraph is central to the inner
  workings of DyNet (\secref{design}), from the user's perspective, the only responsibility is
  to create a new computation graph for each training example.

\end{description}

The overall flow of implementing and training a model in DyNet can be described as follows:
\begin{enumerate}
\item Create a \texttt{Model}.
\item Add the necessary \texttt{Parameter}s and \texttt{LookupParameter}s to the model
\item Create a \texttt{Trainer} object and associate it with the Model.
\item For each example:
\begin{enumerate}
\item Create a new \texttt{ComputationGraph}, and populate it by building an
  \texttt{Expression} representing the desired computation for this example.
\item Calculate the result of that computation forward through the graph by calling the \texttt{value()} or \texttt{npvalue()} functions of the final \texttt{Expression}
\item If training, calculate an \texttt{Expression} representing the loss function, and use it's \texttt{backward()} function to perform back-propagation
\item Use the \texttt{Trainer} to update the parameters in the \texttt{Model}
\end{enumerate}
\end{enumerate}

The contrast with static declaration libraries such as Theano and TensorFlow can be found in the fact that the ``create a graph'' step falls within our loop over each example.
This has the advantage of allowing the user to flexibly create a new graph structure for each instance and to use flow control syntax (e.g., iteration) from their native programming language to do so.
Of course, it also adds the requirement that graph construction be fast enough that it does not present a burden, a challenge which we address in \secref{design}.

\subsection{High-level Example}

\begin{figure}
  \centering
  \begin{minted}[mathescape,
                 linenos,
                 numbersep=5pt,
                 gobble=2,
                 frame=lines,
                 framesep=2mm]{python}
  import dynet as dy
  import numpy as np
  model = dy.model()
  W_p = model.add_parameters((20, 100))
  b_p = model.add_parameters(20)
  E   = model.add_lookup_parameters((20000, 50))
  trainer = dy.SimpleSGDTrainer(model)
  for epoch in range(num_epochs):
    for in_words, out_label in training_data:
      dy.renew_cg()
      W = dy.parameter(W_p)
      b = dy.parameter(b_p)
      score_sym = dy.softmax(W*dy.concatenate([E[in_words[0]],E[in_words[1]]])+b)
      loss_sym = dy.pickneglogsoftmax(score_sym, out_label)
      loss_val = loss_sym.value()
      loss_sym.backward()
      trainer.update()
    correct_answers = 0
    for in_words, out_label in test_data:
      dy.renew_cg()
      W = dy.parameter(W_p)
      b = dy.parameter(b_p)
      score_sym = dy.softmax(W*dy.concatenate([E[in_words[0]],E[in_words[1]]])+b)
      score_val = score_sym.npvalue()
      if out_label == np.argmax(score_val):
        correct_answers += 1
    print(correct_answers/len(test_data))
  \end{minted}
  \caption{An example of training and testing in the DyNet Python API.}
  \label{fig:outline}
\end{figure}

To illustrate DyNet's coding paradigm on a high level, we demonstrate a sketch of a DyNet programs in Python Figure~\ref{fig:outline}.
This program shows the process of performing maximum likelihood training for a
simple classifier that calculates a vector of scores for each class it will be
expected to predict, then returns the ID of the class with the highest score.
We assume each training example to be an \texttt{(input, output)} pair, where
\texttt{input} is tuple of two word indices, and \texttt{output} is a number
indicating the correct class.

In the first two lines we import the appropriate libraries.
In Line 3, we initialize the DyNet model, which allocates space in memory to hold parameters, but does not initialize them.
In Lines 4--6, we add our parameters to the model; this process will be different
depending on the specific variety of model we will be using. Here, we add a
$20\times 100$ weight matrix, a $20$-dim bias vector, and a lookup-table
(embedding table) mapping a vocabulary of $20,000$ items to $50$-dim vectors.
In Line 7, we initialize a trainer (in this case a simple stochastic gradient descent (SGD) trainer), which will be in charge of updating the parameters in the model.
We then begin multiple epochs of training and testing over the data in Line 8.

Starting at Line 9, we iterate over the training data.
Line 10 clears the current computation graph, starting a blank graph for the
current computation.
In Lines 11--13, we create a graph that will calculate a score vector for the
training instance (this process is also model-dependent). Here, we first access the
weight matrix and bias vector parameters living in the model (\texttt{W\_p} and \texttt{b\_p}),
and add them to the graph as expressions (\texttt{W} and \texttt{b\_p}) for use in this
particular training example. We then lookup the
two vectors corresponding to the input ids, concatenate them, do a linear
transform followed by a softmax, creating an expression corresponding to
computation.
Then, in line 14 we create an expression
corresponding to the loss --- the negative log likelihood of the correct answer
after taking a softmax over the scores.
In Line 15, we calculate the value of the computation by performing computation forward through the graph, and in Line 16 we perform the backward pass, accumulating the gradients of the parameters in the ``model'' variable.
In line 17, we update the parameters according to the SGD update rule, and clear the accumulated gradients from the model.

Next, starting on Lines 18 and 19, we step through the testing data and measure accuracy.
In Lines 20--23 we again clear the computation graph and define the expression
calculating the scores of the test data in the same manner that we did during training.
In Line 24 we perform the actual calculation and retrieve the scores as a NumPy array from the graph.
In Lines 25 and 26, we determine whether the correct label is the one with the highest score, and count it as a correct answer if so.
Finally in Line 27, we print the test accuracy for this iteration.

\subsection{Two Examples of Dynamic Graph Construction}
\label{sec:dynamicexamples}

In this section, we demonstrate two examples: one of a dynamic network where the structure of the network changes for each training example, and another where we perform dynamic flow control based on the results of computation.

\begin{figure}
  \centering
  \begin{minted}[mathescape,
                 linenos,
                 numbersep=5pt,
                 gobble=2,
                 frame=lines,
                 framesep=2mm]{python}
      class TreeRNNBuilder(object):
          def __init__(self, model, word_vocab, hdim):
              self.W = model.add_parameters((hdim, 2*hdim))
              self.E = model.add_lookup_parameters((len(word_vocab),hdim))
              self.w2i = word_vocab

          def encode(self, tree):
              if tree.isleaf():
                return self.E[self.w2i.get(tree.label,0)]
              elif len(tree.children) == 1: # unary node, skip
                expr = self.encode(tree.children[0])
                return expr
              else:
                assert(len(tree.children) == 2)
                e1 = self.encode(tree.children[0])
                e2 = self.encode(tree.children[1])
                W = dy.parameter(self.W)
                expr = dy.tanh(W*dy.concatenate([e1,e2]))
                return expr

      model = dy.Model()
      U_p = model.add_parameters((2,50))
      tree_builder = TreeRNNBuilder(model, word_vocabulary, 50)
      trainer = dy.AdamTrainer(model)
      for epoch in xrange(10):
        for in_tree, out_label in read_examples():
          dy.renew_cg()
          U = dy.parameter(U_p)
          loss = dy.pickneglogsoftmax(U*tree_builder.encode(in_tree), out_label)
          loss.forward()
          loss.backward()
          trainer.update()
  \end{minted}
 \caption{An example of tree-structured recursive neural network.}
  \label{fig:treenn}
\end{figure}

\textbf{Dynamic Network Shape:}
In Figure \ref{fig:treenn} we provide an example for a tree-shaped recursive
neural network, following the model of \cite{socher11recursivenn}. Each example is a tree with
unary or binary nodes, and leaves that come from a vocabulary $V$. The tree is
recursively encoded as a vector according to the following rules: the encoding
of a leaf is the embedding vector corresponding to the leaf's vocabulary item.
The encoding of a unary node is the same as the encoding of its child node, and
the encoding of a binary node $n$ with child nodes $c_1$ and $c_2$ is a linear
combination of the encoding of the nodes followed by a $\tanh$ activation
function: $\textit{enc}(n) = \tanh(W*[\textit{enc}(c_1);\textit{enc}(c_2)])$ where $W$ is a model parameter
and $[;]$ denotes vector concatenation. Computing the tree encoding is performed
using a neural network that is dynamically created based on the tree structure.
This is handled by the \texttt{TreeRNNBuilder} class (lines 1--19), to be discussed shortly.
Once the expression representing the tree is available, we then use it for
classification by a linear transformation followed by a softmax (line 28).

The \texttt{TreeRNNBuilder} class holds the model parameters (the transformation
parameter $W$ and the leaf-word embeddings $E$). These are initialized in lines
2--5. \texttt{word\_vocab} is a table mapping vocabulary items to numeric IDs
(indices in $E$).
The encoding itself is done in the recursive function \texttt{encode(self, tree)}
(lines 6--19). This function takes as input a tree data-structure (implementation
not provided) that has the basic tree operations. It then follows the tree
encoding logic: if the tree is a leaf, return the expression of the vector corresponding
to the leaf's word (\texttt{tree.label}). If the tree is a unary tree,
recursively call the encoding function of its child. Finally, if the tree is a
binary tree, recursively compute the encoding of each child (lines 15 and 16), resulting in two
expression vectors that are then combined into the expression of the current
tree (line 18).

The tree encoder is driven by the main program. For each training instance, we
create a new computation graph, and then encode the tree as a vector, multiply
the resulting encoding by $U$, pass through a softmax function, compute the
loss, compute the gradients by calling backwards, and update the parameters.

Notably, the code for computing the tree-structured network is straightforward
and short (19 lines of code), and extending it with a more complex composition function such as the
Tree LSTM of Tai et al \cite{tai15treelstm} is straightforward (the corresponding
class for the TreeLSTM code that we use in the benchmarks is 39 lines of
readable Python code).

\begin{figure}
  \centering
  \begin{minted}[mathescape,
                 linenos,
                 numbersep=5pt,
                 gobble=2,
                 frame=lines,
                 framesep=2mm]{python}
  import dynet as dy
  import numpy as np
  model = dy.model()
  W_p = model.add_parameters(50)
  b_p = model.add_parameters(1)
  E   = model.add_lookup_parameters((20000, 50))
  trainer = dy.SimpleSGDTrainer(model)
  for epoch in range(num_epochs):
    for in_words, out_label in training_data:
      dy.renew_cg()
      W = dy.parameter(W_p)
      b = dy.parameter(b_p)
      score_sym = W*sum([ E[word] for word in in_words ]) + b
      loss_sym = dy.logistic( out_label*score )
      loss_sym.backward()
      trainer.update()
    correct_answers = 0
    for in_words, out_label in test_data:
      dy.renew_cg()
      W = dy.parameter(W_p)
      b = dy.parameter(b_p)
      score_sym = b
      for word in in_words:
        score_sym = score_sym + W * E[word]
        if abs(score_sym.value()) > threshold:
          break
      if out_label * score_sym.value() > 0:
        correct_answers += 1
    print(correct_answers/len(test_data))
  \end{minted}
  \caption{An example of dynamic flow control.}
  \label{fig:dynamicexample}
\end{figure}

\textbf{Dynamic Flow Control:}
There are a number of reasons why we would want to perform flow control based on the results of neural network computation.
For example, we may want to search for the best hypothesis in a model where finding the exact best answer is inefficient, a common problem in models that rely on sequences of actions, such as selections of words in a translation \cite{sutskever14sequencetosequence}, shift-reduce actions to generate a parse tree \cite{dyer2015stacklstm}, or selection of component networks to perform question answering \cite{andreas2016compose}.

In \figref{dynamicexample}, we show a simplified example of such a situation: a binary text classifier that at test time can make a decision before reading the whole document, similar to the test-time behavior of the model proposed by \cite{iyyer2014factoid}.
This allows the classifier to avoid wasting processing time when the answer is clear, or to answer quickly in a setting when we get words in real-time, such as from a speech recognition system.
The first 16 lines show a training algorithm very similar to the simpler text classifier in \figref{outline}, with the exception that we are now using the sum of the word embeddings for all the words (Line 13), and only predicting a $\{-1,1\}$ binary label using the logistic sigmoid function (Line 14).
The dynamic flow control specifically comes into play in the test-time inference algorithm in Lines 22--26.
In Line 22, we initialize our estimate score with the value of the bias parameter, and starting at Line 33 we loop over the input.
In Line 24, we update our current estimate of the score with the contribution of the next word by multiplying its embedding with the weight vector.
In Lines 25 and 26, we check whether the absolute value of the current score is above some pre-defined threshold, and if it is, terminate the loop and output our predicted label.

It should be noted that here we are only performing dynamic flow control at test time.
While it is also possible to perform dynamic flow control at training time, this requires more sophisticated training algorithms using reinforcement learning.
These algorithms, require interleaving model evaluation and decision making on the basis of that evaluation, are well-suited to implementation within DyNet, but they are beyond the scope of this introductory paper.

%% file: 030-design.tex

As detailed in the previous section, one major feature that sets DyNet apart from other neural network toolkits is its ability to efficiently create new computation graphs for every training example or minibatch.
To maintain computational efficiency, DyNet uses careful memory management to store values associated with forward and backward computation (\secref{overhead}), so the majority of time can be spent on the actual computation (\secref{computation}).

\subsection{Computation Graphs}
\label{sec:computation graphs}

Behind the scenes, DyNet maintains a \texttt{ComputationGraph} which is a directed acyclic graph composed of \texttt{Node} objects.
A node instance represents a variable with a certain shape containing parameters, constants, input data, and, most commonly, the result of applying a single elementary function to the node's inputs (the shape of the node is inferred when the node object is created based on the shapes of the node's inputs). Each node maintains a list of incoming edges stored as an ordered list of references to other nodes which represent the inputs to the function computed by the node (nodes representing parameters, constants, and random variate generators have no inputs). 
DyNet defines nodes for computing a wide range of elementary functions, including common ones such as addition, multiplication, softmax, tanh, item selection, as well as many others.
\texttt{Expression} objects are thin wrappers around nodes, which abstract away from this behind-the-scenes detail of the computation graph structure and behave syntactically like a tensor value.

Each node has forward and backward functions specifically implemented by the DyNet library.
Forward computation takes the inputs $x_1,x_2,\ldots$ and calculates the function $f(x_1,x_2,\ldots)$ corresponding to the operation that the node represents.
Backward computation takes inputs $x_1,x_2,\ldots$, result $f(x_1,x_2,\ldots)$, the derivative of the loss $L$ with respect to the function $\frac{dL}{df(x_1,x_2,\ldots)}$, and a variable index $i$, and it returns the derivative of variable $x_i$ with respect to the loss, $\frac{dL}{x_i}$.
Thus, if a user wants to implement a new operation not supported by DyNet, they will need to implement these two functions (along with several utility functions to define the corresponding expression wrappers, etc.).

\newsavebox{\mintedbox}
\begin{lrbox}{\mintedbox}
\begin{minipage}{0.6\textwidth} 
\vspace{-3.5cm}
\begin{minted}[mathescape,
               linenos,
               numbersep=5pt,
               gobble=0,
               frame=lines,
               framesep=2mm]{python}
b = dy.parameter(cg, b_param)
W1 = dy.parameter(cg, W1_param)
W2 = dy.parameter(cg, W2_param)
x = dy.vectorInput(x_vec)
e = e_param[i]
g = dy.tanh(W1 * x + b) + dy.tanh(W2 * e + b)
g_val = g.value()
g.backward()
\end{minted}
\end{minipage}
\end{lrbox}

\begin{figure*}[t!]
  \centering
  \subfigure[Computation graph]{
    \includegraphics[width=0.35\textwidth]{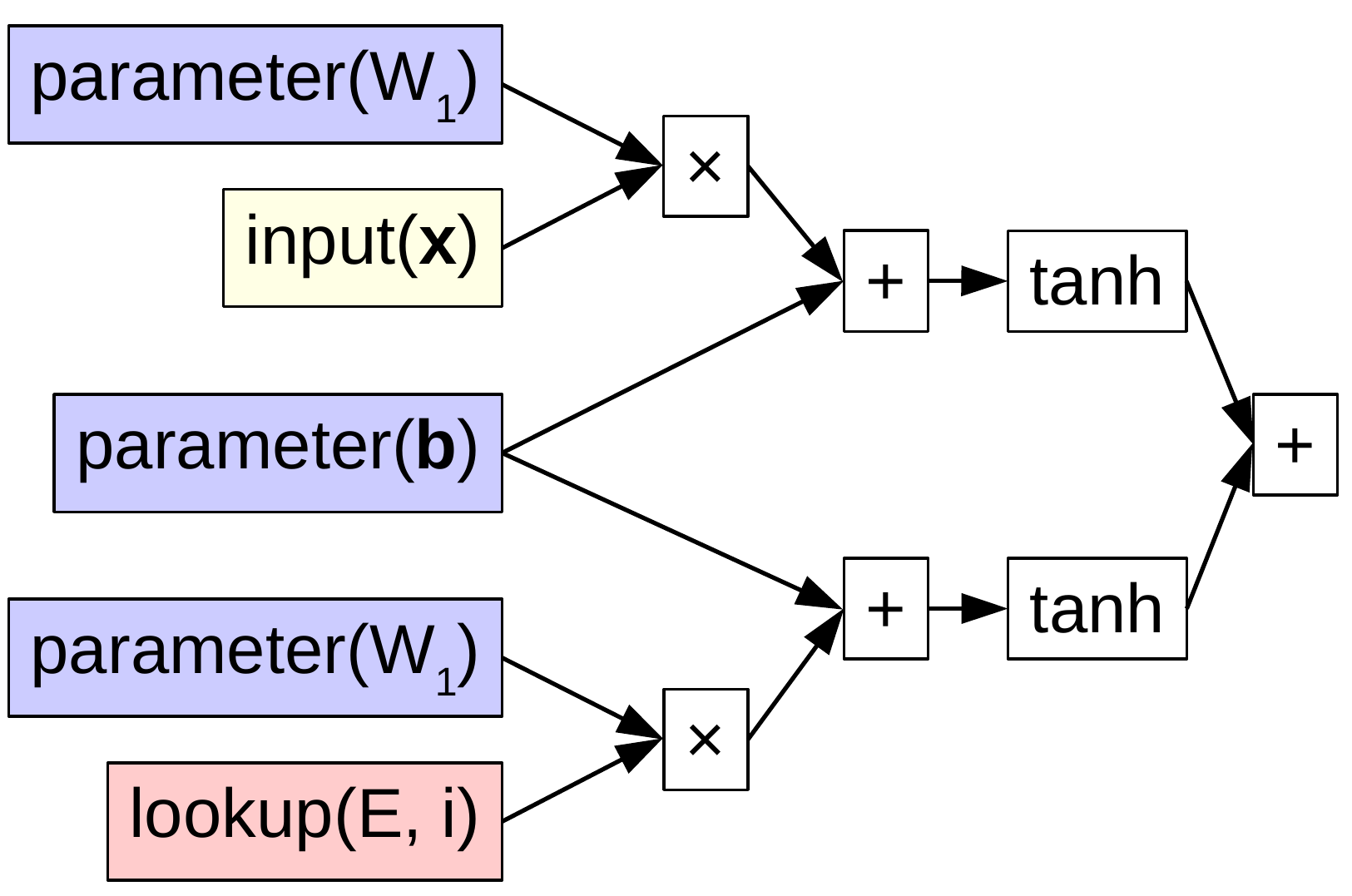}
  }%
  ~
  \subfigure[Corresponding code]{
    \usebox{\mintedbox}
  }
\caption{An example of a computation graph for $g(\boldsymbol{x},j) = \text{tanh}(W_1*\boldsymbol{x}+\boldsymbol{b}) + \text{tanh}(W_2*e_j+\boldsymbol{b})$, and the corresponding code.}
\label{fig:computationgraph}
\end{figure*}

In \figref{computationgraph}, we show a computation graph for a computation consisting standard and lookup parameters, a constant input, parameter nodes, lookup parameter nodes, and several elementary functions.
Lines 1-6 of the above code will perform symbolic computation to create the computation graph, which will set up a data structure consisting of nodes and pointers to their inputs.
Line 7 sequentially steps through each node (in the order of addition to the graph), and for each node assigns a pointer to memory to hold the calculation results, then calls the ``forward'' function to populate this memory with the calculation result.%
\footnote{This is in contrast to Chainer, another dynamic toolkit, which performs the forward step while it is performing the computation graph construction. One advantage of the DyNet design of method of performing symbolic computation first is that theoretically it would be possible to create graph optimization routines that run before performing actual computation. We discuss these shortly as future work in \secref{conclusion}.}
Line 8 then sequentially starts at the designated final node in the graph, steps backward through each node and, if it has at least one input node, performs back-propagation of the derivatives of the loss.
For nodes corresponding to parameters, gradients are also accumulated to later be used by the trainer in performing parameter updates.


\subsection{Efficient Graph Construction}
\label{sec:overhead}

DyNet's backend is designed with this rapid computation graph construction in mind.
First and foremost, the backend is written in C++ with a specific focus on ensuring that graph-building operations are as efficient as possible.

Specifically, one of the major features of DyNet is that it performs its own memory management for all computation results.%
\footnote{The data structure used to represent the computation graph is managed by the standard C++ memory management, care is taken to avoid performing expensive allocations in the construction of this structure as well.}
When the DyNet library is initialized, it allocates three large blocks of memory, the one for storing results of forward calculation through the graph, one for storing backward calculations, and one for storing parameters and the corresponding gradients.
When starting processing of a new graph, the pointers to the segments of memory responsible for forward and backward computation are reset to the top of the corresponding segments, and any previously used backward memory is zeroed out with a single operation.\footnote{This discrepancy is due to the fact that memory owned by a particular node is affected by a single operation in the forward step, but the backward step may have to accumulate gradients from multiple operations. For example, the $b$ node in \figref{computationgraph} is used in multiple operations, and thus must receive gradients passed back from both of these operations.}
When more memory is required during forward execution or in advance of running the backward algorithm, the custom allocator simply returns the current pointer, then increments it according to the size of the memory segment requested.
Thus, in this extremely simple memory allocation setup, deallocation and allocation of memory are simple integer arithmetic.
This is also an advantage when performing operations on the GPU, as managing memory requires no interaction with the GPU itself, and can be handled entirely by managing, on the CPU, pointers into GPU memory.

Interacting with the DyNet library is also efficient.
When writing programs in C++, it is possible to interact directly on the machine code level. The Python wrapper is also written with efficiency in mind, attempting to keep
the wrapping code as minimal as possible, delegating the operations directly to the C++ core when possible, writing the glue code in Cython rather than Python, and trying to minimize Python objects allocation. At the same time, we also
attempt to provide a ``Pythonic'' and natural high-level API.
We demonstrate empirically in \secref{evalresults} that the Python speed is indeed very close to the C++ one.

\subsection{Performing Computation}
\label{sec:computation}

While computation graph construction is particularly important given DyNet's focus on dynamic declaration, like any other neural network toolkit it is also paramount that the actual computation be efficient.
In order to achieve efficient execution of arithmetic operations, DyNet relies on the Eigen \cite{guennebaud2010eigen} library, which is also used to provide the backend for TensorFlow.
Eigen is a general-purpose matrix/tensor math library that is optimized for speed, allows for flexible execution and combination of operations using C++ templates, and allows for simple portability between CPU and GPU.

Most of the operations provided by DyNet are implemented in pure Eigen, but in a few cases DyNet implements its own custom operations to improve efficiency, or to provide functionality not currently available in Eigen itself.
In addition, DyNet allows for a number of methods to further improve efficiency by exploiting parallelism, and also wrappers around common high-level functionalities to ease programming, as described in the next section.

%% file: 040-abstractions.tex
As mentioned in \secref{codingparadigm}, DyNet implements operations that represent elementary (sub-)differentiable functions over tensors.
This is similar to the operations provided in libraries such as Theano or TensorFlow.

\begin{figure}
 \centering
 \includegraphics[width=15cm]{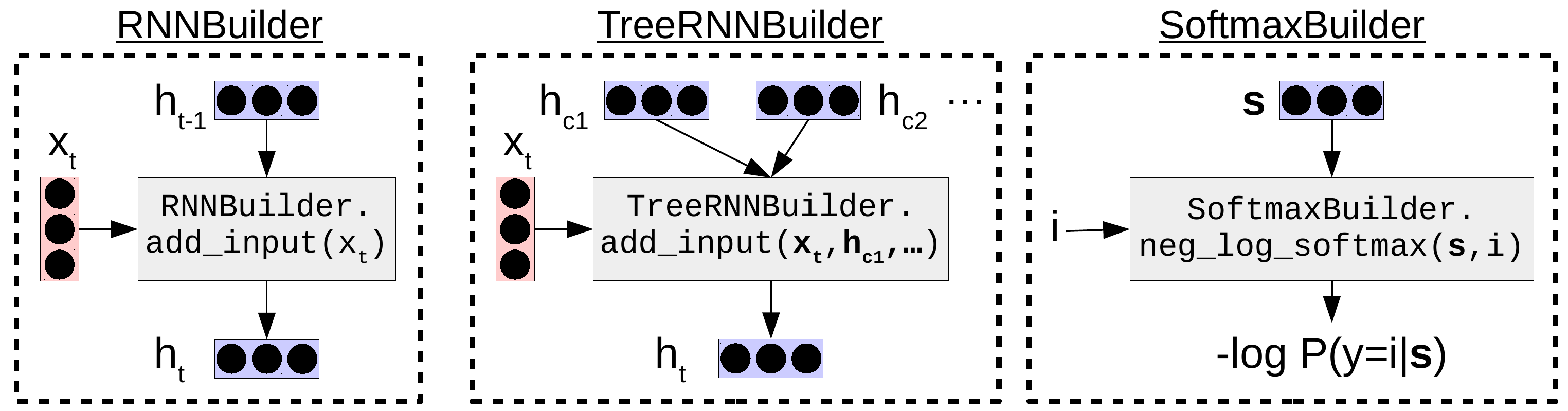}
 \caption{An example of higher-level constructs implemented as DyNet \texttt{Builder}s, along with their canonical usage.}
 \label{fig:builders}
\end{figure}

In addition to these elementary operations, it is also common to use more complicated structures that can be viewed as combinations of these elementary operations.
Common examples of this include recurrent neural networks (RNNs), tree-structured networks, and more complicated methods for calculating softmax probability distributions.
In other libraries, these higher-level constructs are either provided natively, or through a third-party library such as Keras.
In DyNet, native support for recurrent networks, tree-structured networks, and more complicated softmax functions is provided through \texttt{Builder}s, which we describe in detail below and summarize in Figure \ref{fig:builders}.

\subsection{Recurrent Neural Network Builders}
\label{sec:rnnbuilder}

The first, and most commonly used, variety of builder in DyNet is builders to build RNNs, \texttt{RNNBuilder}s.
The \texttt{RNNBuilder} parent class describes the overall interface, and concrete implementations are provided by child classes.
DyNet implements builders to implement simple Elman-style recurrent networks (\texttt{SimpleRNNBuilder}; \cite{elman90rnn}), as well as long short-term memory (\texttt{LSTMBuilder}; \cite{hochreiter97lstm}) or gated recurrent unit (\texttt{GRUBuilder}; \cite{chung2014gru}).

The RNN interface's canonical usage is as follows:
\begin{enumerate}
\item At the beginning of training, calling the constructor of some variety of \texttt{RNNBuilder}, will add the parameters to the DyNet model. The size and number of hidden layers in the RNN are specified as parameters.
\item At the beginning of the sequence to be processed, start a new sequence and get the initial state of the RNN.
\item Every time an input is received, we call \texttt{add\_input(x)}, where \texttt{x} is the input, then get the output for further processing.
\end{enumerate}
This is demonstrated in code in \figref{rnnbuilder}(a).

\begin{figure}
 \centering
    \begin{minted}[mathescape,
                   linenos,
                   numbersep=5pt,
                   gobble=2,
                   frame=lines,
                   framesep=2mm]{python}
    RNN = dy.LSTMBuilder(LAYERS, EMBED, HIDDEN, model)     # Initialization
    for example in training_data:
      state = RNN.initial_state()                          # Starting sequence
      for x in example:
        state = state.add_input(x)                         # RNN update
        do_further_processing(state.output())              # Processing
    \end{minted}
    (a) Canonical usage of RNNBuilder in DyNet.
    \begin{minted}[mathescape,
                   linenos,
                   numbersep=5pt,
                   gobble=2,
                   frame=lines,
                   framesep=2mm]{python}
    # Graph definition
    cell = tf.nn.rnn_cell.BasicLSTMCell(HIDDEN) 
    cell = tf.nn.rnn_cell.MultiRNNCell([cell] * LAYERS)
    outputs, _ = tf.nn.dynamic_rnn(cell, x_input, sequence_length=x_len)
    x_result = do_further_processing(outputs)
    # Execution
    with tf.Session() as sess:
      for example in training_data:
        result = sess.run(x_result, feed_dict={x_input: x_in, x_len: len(x_in)})
    \end{minted}
    (b) Similar code in TensorFlow.
    \begin{minted}[mathescape,
                   linenos,
                   numbersep=5pt,
                   gobble=2,
                   frame=lines,
                   framesep=2mm]{python}
    RNN = dy.LSTMBuilder(LAYERS, EMBED, HIDDEN, model)     # Initialization
    for example in training_data:
      outputs = RNN.initial_state.transduce(example)       # Transduction
      for h in outputs:
        do_further_processing(h)                           # Processing
    \end{minted}
    (c) The DyNet sequence processing interface.
 \caption{Various RNN interfaces.}
 \label{fig:rnnbuilder}
\end{figure}

This is in contrast to the standard practice in the static declaration paradigm, which generally consists of creating a full string of inputs, then calling a specialized function that runs the RNN over all of the inputs in this sequence.
An example of this in TensorFlow is shown in \figref{rnnbuilder}(b), where at the graph definition stage we define a recurrent neural network, and call a specialized function \texttt{dynamic\_rnn()} that runs the RNN over a full sequence.

This method of processing a full sequence at a time has advantages and disadvantages.
A first advantage is that once the user is familiar with the API and know how to shape their input properly, they can process a full sequence with a single function call, as opposed to writing a for loop.
A second advantage is that (as explained in more detail \secref{minibatching}) some computations can be shared across multiple inputs, improving efficiency.
However, one major disadvantage is that this method requires that the full sequence must be homogeneous, and known before the processing begins.
This makes things difficult when, for example, we would want to interleave RNN processing with other processes that may differ based on the structure of the input or based on the predictions of the RNN.

Because of this, DyNet also supports a sequence processing API\footnote{With parts in the master branch, and other optimized functions residing in the \texttt{sequence-ops} branch as of this writing.} as shown in \figref{rnnbuilder}(c), which allows a user to call functions that look up embeddings, transduce, and calculate loss for whole sequences in a single function call.
This allows users who are experienced with the API to reduce the number of function calls and achieve some efficiency gains in cases where this is convenient, while still allowing for usage of the more flexible (and arguably intuitive) canonical API.

\subsection{Tree-structured Neural Network Builders}
\label{sec:treernnbuilder}

In addition, DyNet has \texttt{Builders} to support the creation of tree-structured neural networks, as shown in the middle of \figref{builders}.
These are similar to recurrent neural networks, but take not only a single input from the past, but compose multiple inputs into an output representation.
As mentioned in \secref{dynamicexamples}, these models are relatively simple to implement in DyNet, taking few dozens of Python code, and resulting in speed which is comparable if not identical to that of a C++ DyNet implementation.
Yet, the built-in builders provide out-of-the-box efficient and debugged implementations of tree-structured networks such as recursive neural networks \cite{socher11recursivenn} or tree-structured LSTMs \cite{tai15treelstm} for users who just want to get started working using these networks as tools.

\subsection{Large-Vocabulary Softmax Builders}
\label{sec:softmaxbuilder}

Finally, DyNet supports a \texttt{SoftmaxBuilder} interface for different methods of calculating softmax probability distributions with large output spaces given an input, as shown on the right side of \figref{builders}.
These builders include class-based or hierarchical softmax \cite{goodman2001classes,mikolov2011extensions}, which can speed training by dividing calculation into multiple steps.
At training time, these methods allow calling the \texttt{neg\_log\_softmax()} function, which rapidly calculates the loss for a softmax function.
At test time, it is possible to either sample outputs from the distribution, or calculate the full probability distribution.

%% file: 045-efficiency.tex
DyNet contains a number of features to improve computational efficiency, including sparse updates, minibatching, and multi-processing across CPUs.

\subsection{Sparse Updates}
\label{sec:sparseupdates}

As mentioned in \secref{codingparadigm}, DyNet supports two types of parameters, \texttt{Parameters} and \texttt{LookupParameters}.
In particular, lookup parameters are designed to perform a sparse lookup of only a few vectors or tensors for each training instance (e.g. only the word embeddings for words that exist in a particular sentence).
Because of this, when performing training, only the gradients of parameters that existed in the training instance that we are currently handling will be non-zero, and the rest will be zero.

In update rules such as the standard stochastic gradient descent update
\begin{equation}
\theta_i \leftarrow \theta_i - \alpha \frac{dL(x)}{d\theta_i}
\end{equation}
that moves parameter $\theta_i$ in a direction that decreases loss $L(x)$ with step size $\alpha$, if gradients are zero, there will be no change made to the parameters.
Nonetheless, in standard implementations, it is common to loop over all the parameters and update them, regardless of whether their gradients are zero or not, which can be wasteful when gradients of the majority of parameters are zero.

DyNet's sparse update functionality takes advantage of this fact by keeping track of the identities of all vectors that have been accessed through a \texttt{lookup} operation, and only updates the parameters that have non-zero gradients.
This can greatly improve speed in cases where we have a large variety of parameters to be accessed by \texttt{lookup} operations, with two caveats.
The first caveat is that while uniform speedups can be expected on CPU, on GPUs (which specialize in rapidly performing large numbers of parallel operations), the amount of time to loop over all parameters can be negligible.
The second caveat is that while sparse updates are strictly correct for many update rules such as the simple SGD above and AdaGrad \cite{duchi2011adaptive}, other update rules such as SGD with momentum \cite{nesterov1983method} or Adam \cite{kingma2014adam} perform updates even on steps where gradients are zero, and will therefore not produce exactly the same results when sparse updates are performed.%
\footnote{While there are computational tricks that perform just-in-time update of parameters to allow both efficient and correct updates \cite{shalev2011pegasos}, in interest of maintaining the simplicity of the programming interface, these are not implemented in DyNet at this time.}

\subsection{Minibatching}
\label{sec:minibatching}

\begin{figure}
 \centering
 \includegraphics[width=13cm]{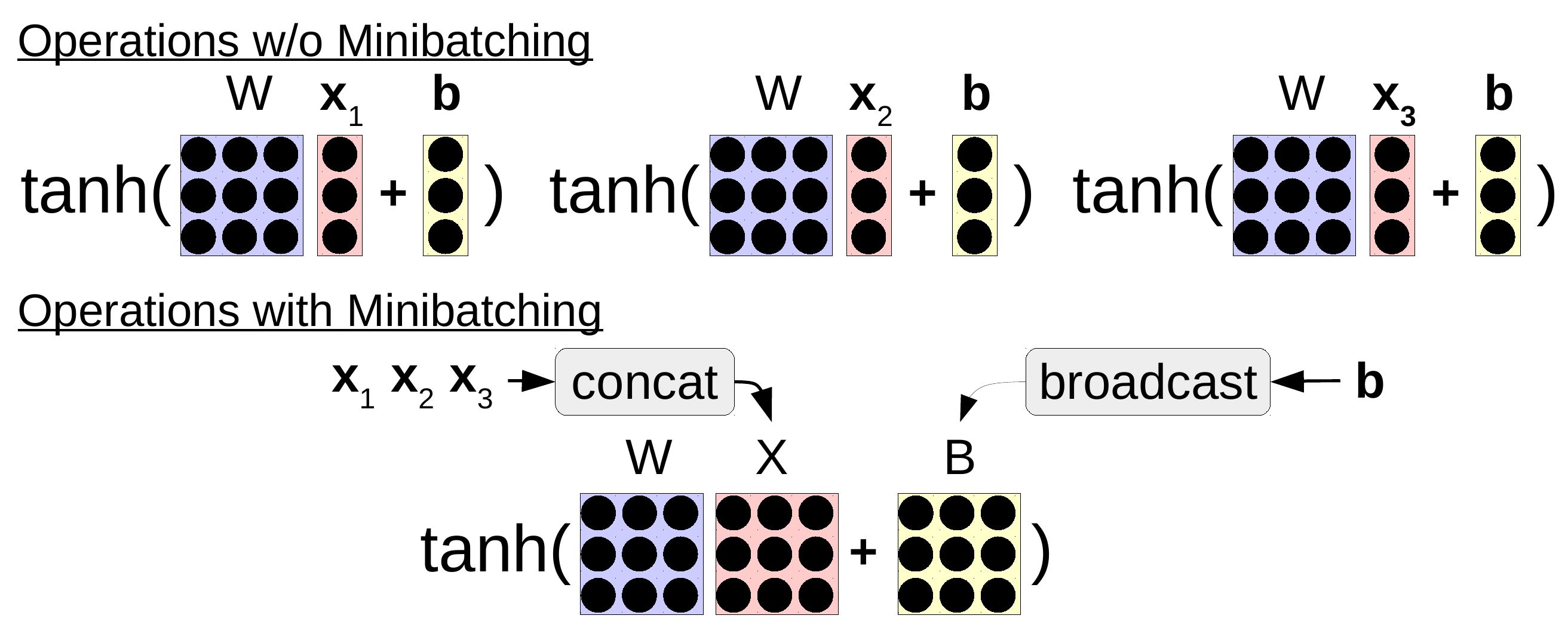}
 \caption{An example of minibatching for an affine transform followed by a tanh nonlinearity.}
 \label{fig:minibatching}
\end{figure}

Minibatching takes multiple training examples and groups them together to be processed simultaneously, often allowing for large gains in computational efficiency due to the fact that modern hardware (particularly GPUs, but also CPUs) have very efficient vector processing instructions that can be exploited with appropriately structured inputs.
As shown in Figure~\ref{fig:minibatching}, common examples of this in neural networks include grouping together matrix-vector multiplies from multiple examples into a single matrix-matrix multiply, or performing an element-wise operation (such as $\tanh$) over multiple vectors at the same time as opposed to processing single vectors individually.

In most neural network toolkits, mini-batching is largely left to the user, with a bit of help from the toolkit.
This is usually done by adding an additional dimension to the tensor that they are interested in processing, and ensuring that all operations consider this dimension when performing processing.
This adds some cognitive load, as the user must keep track of this extra batch dimension in all their calculations, and also ensure that they use the correct ordering of the batch dimensions to achieve maximum computational efficiency.
Users must also be careful when performing operations that combine batched and unbatched elements (such as batched hidden states of a neural network and unbatched parameter matrices or vectors), in which case they must concatenate vectors into batches, or ``broadcast'' the unbatched element, duplicating it along the batch dimension to ensure that there are no illegal dimension mismatches.

\begin{figure}
 \centering
  \begin{minted}[mathescape,
                linenos,
                numbersep=5pt,
                gobble=2,
                frame=lines,
                framesep=2mm]{python}
  # in_words is a tuple (word_1, word_2)
  # out_label is an output label
  word_1 = E[in_words[0]]
  word_2 = E[in_words[1]]
  scores_sym = dy.softmax(W*dy.concatenate([word_1, word_2])+b)
  loss_sym = dy.pickneglogsoftmax(scores_sym, out_label)
  \end{minted}
  (a) Non-minibatched classification.
  \begin{minted}[mathescape,
                linenos,
                numbersep=5pt,
                gobble=2,
                frame=lines,
                framesep=2mm]{python}
  # in_words is a list [(word_{1,1}, word_{1,2}), (word_{2,1}, word_{2,2}), ...]
  # out_labels is a list of output labels [label_1, label_2, ...]
  word_1_batch = dy.lookup_batch(E, [x[0] for x in in_words])
  word_2_batch = dy.lookup_batch(E, [x[1] for x in in_words])
  scores_sym = dy.softmax(W*dy.concatenate([word_1_batch, word_2_batch])+b)
  loss_sym = dy.sum_batches( dy.pickneglogsoftmax_batch(scores_sym, out_labels) )
  \end{minted}
  (b) Minibatched classification.
 \caption{A contrast of non-minibatched and mini-batched classifiers. The only differences are the necessity to input multiple word IDs, and calculate loss over multiple labels.}
 \label{fig:minibatchingcbow}
\end{figure}

DyNet hides much of this complexity from the user through the use of specially designed batching operations which treat the number of mini-batch elements not as another standard dimension, but as a special dimension with particular semantics.
Broadcasting is done behind the scenes by each operation implemented in DyNet, and thus the user must only think about inputting multiple pieces of data for each batch, and calculating losses using multiple labels.
An example of this is shown in Figure~\ref{fig:minibatchingcbow}.
This example implements the Lines 13 and 14 of the text classifier in \figref{outline}, which performs the main computation of the loss function for training using batches.
We can see there are only 4 major changes: the word IDs need to be transformed into lists of IDs instead of a single ID, we need to call \texttt{lookup\_batch} instead of the standard lookup, we need to call \texttt{pickneglogsoftmax\_batch} instead of the unbatched version, and we need to call \texttt{sum\_batches} at the end to sum the loss from all the batches.%
\footnote{In the case of minibatching sequences, it is necessary perform \texttt{lookup\_batch} and \texttt{pickneglogsoftmax\_batch} in lock-step across identical time-steps in multiple sentences. Example code for how to do so is included in the benchmarks in \secref{eval}, but a detailed description is beyond the scope of this paper. Minibatching across more complex structures such as trees requires more complex algorithms \cite{looks2017dynamic}, and integrating these into DyNet is an interesting challenge for the future.}

\subsection{Parallel Processing}
\label{sec:parallelization}
 
In addition to minibatch support, DyNet also supports training models using many CPU cores.
Again, DyNet abstracts most of the behind-the-scenes grit from the user.
The user defines a function to be called for each datum in the training data set, and passes this function, along with an array of data, to DyNet.
Internally, DyNet launches a pool of training processes and automatically handles passing data examples to each worker.
Each worker process individually processes a datum, computing the results of the forward and backward passes, computes gradients with respect to each parameter, and passes these results back to the parent process via a shared memory variable.
Whenever the parent process, which is also processing data, completes a gradient computation, it averages all of the gradients currently in the shared memory gradient storage and updates all parameters with respect to that average gradient.
In this way running training on $n$ cores is similar to training with a stochastic minibatch size with expected value $\approx n$.
This method is quite efficient, achieving nearly perfectly linear speedups with increasing numbers of cores, due to its lockless nature.
Though lockless optimization methods compute only approximate gradients due to overlapping parameter reads and writes, in practice they show virtually no degradation in performance versus fully serial or lock-based methods \cite{recht2011hogwild,sun2016asynchronous}.

%% file: 050-eval.tex
In this section, we compare the C++ and Python interfaces to DyNet to three other popular libraries: Theano \cite{bergstra2010theano}, TensorFlow \cite{abadi2016tensorflow}, and Chainer \cite{tokui2015chainer}.\footnote{
  DyNet used the Jan 4 09:02:17 of the \texttt{sequence-ops} branch, Theano used version 0.8.2, TensorFlow used version 0.12.0, and Chainer used version 1.19.0.
}
We choose these\footnote{Among the myriad of others such as Torch \cite{collobert2002torch}, MxNet \cite{chen2015mxnet}, and CNTK \cite{yu2014cntk}.} because Theano and TensorFlow are currently arguably the most popular deep learning libraries, and because Chainer's define-by-run philosophy is similar to DyNet's.

\subsection{Tasks}
\label{sec:evaltasks}

We evaluate DyNet on 4 natural language processing tasks.
Code implementing these benchmarks is available at \url{http://github.com/neulab/dynet-benchmark}.
\begin{description}
\item[Recurrent Neural Network Language Model:]
Similarly to \cite{sundermeyer12lstmlm}, we train a recurrent neural network language model using LSTM units.
This is arguably one of the most commonly, and also most simply structured tasks in natural language processing.
Because of this simplicity it is relatively simple to implement in the static declaration paradigm, and due to its wide recognition most toolkits have been explicitly tested on this task, and arguably will be able to perform at the top of their ability, making it an ideal stress-test.
It is also conducive to mini-batching, so we examine models using mini-batching as well, set to 16 sentences unless otherwise specified.
Models are trained and tested on Tomas Mikolov's version of the Penn Treebank,\footnote{In the ``Basic Examples'' download at \url{http://www.fit.vutbr.cz/~imikolov/rnnlm/}.} using the standard tokenization and specification of unknown words, and measured using dev-set perplexity.
\item[Bi-directional LSTM Named Entity Tagger:]
Similarly to the BI-LSTM model of Huang et al. \cite{huang2015bidirectional}, we train a tagger that uses a bi-directional LSTM to extract features from the input sentence, which are then passed through a multi-layer perceptron to predict the tag of the word.
This is also a relatively simple task to implement in all toolkits described above.
Models were trained and tested on the WikiNER English Corpus \cite{nothman2012wikiner}, and all words with frequency less than five were treated as unknowns.
Accuracy is measured using tag-by-tag accuracy.
\item[Bi-directional LSTM Tagger w/ Optional Character Features:]
We create a tagger identical to the one above with one change: words that have a frequency of at least five use an embedding specifically for that word, and less frequent words use an embedding calculated by running a bi-directional LSTM over the characters in the word and concatenating the resulting representation into a word embedding.
This model has the potential to allow for better generalization, as it can use the spelling of low-frequency words to improve generalization.
However, it is also a model that is more difficult to implement efficiently in the static declaration paradigm, as it has a switch to trade between looking up word embeddings or performing character-based embedding combination, making the flow control more difficult than the standard BiLSTM tagger.
\item[Tree-structured LSTM Sentiment Analyzer:]
Finally, we create a sentiment analyzer based on tree-structured LSTMs \cite{tai15treelstm}.
This model has a complicated structure that changes based on the tree in every sentence, and thus is quite difficult to implement in the static declaration architecture.
Thus, we compare solely with Chainer, which is the only toolkit of the three baselines that allows for simple implementation of these models.
Accuracy is measured using fine-grained sentiment tagging accuracy at the root of the tree.
\end{description}

\begin{table*}[t]
\begin{center}
\begin{tabular}{l|rrrr}
Model               & RNNLM   & BiLSTM Tag    & BiLSTM Tag+Char      & Tree LSTM \\ \hline
Train Sentences     & 42.1k   & 142k          & 142k                 & 8.54k     \\
Train Words         & 888k    & 3.50M         & 3.50M                & 164k      \\
Input Vocabulary    & 10k     & 69.5k         & 30.6k word, 887 char & 18.3k     \\
Output Vocabulary   & 10k     & 9             & 9                    & 5         \\ \hline
Word Embedding      & 128     & 128           & 128                  & 128       \\
Word LSTM Nodes     & 256     & 50            & 50                   & 128       \\
Perceptron Nodes    & -       & 32            & 32                   & -         \\
Char Embedding      & -       & -             & 20                   & -         \\
\end{tabular}
\end{center}
\caption{\label{tab:data} Data and default configurations used for each task}
\end{table*}

Various statistics for each task are shown in Table~\ref{tab:data}.
All models are optimized using Adam \cite{kingma2014adam} with a learning rate of 0.001, and trained until the dev accuracy peaks (in cases where we measure accuracy), or for 10 minutes (in cases where we simply measure processing speed).
Default net sizes were chosen somewhat arbitrarily, although they are similar to those in the original papers, and the respective models achieve reasonable accuracies.
\footnote{Note that we are not aiming to achieve state-of-the-art accuracy, but want to ensure that we are using reasonable settings for our speed comparison.}

\subsection{Evaluation Metrics}
\label{sec:evalmetrics}

We compare based on several evaluation measures that are desirable in our toolkits:
\begin{description}
\item[Computation Speed:]
We measure computation speed in wall-clock words/sec or sentences/sec on both CPU (single-thread Intel Xeon E5-2686 v4) and GPU (NVIDIA Tesla K80).
To ensure that no quirks or issues due to busy machines result in unrealistic speed numbers, we perform three runs of training for each model and take the fastest of the three runs.
\item[Startup Time:]
As Theano and TensorFlow require a graph compilation step before performing any computation, we also measure the amount of time required from starting the program to starting processing of the first training instance (including time to read training data).
This time can be amortized over training instances, but does play a role in the development process, requiring developers to wait between when they modify their implementation and when they begin to view results.
\item[Characters of Code:]
To give a very rough estimate of the amount of work that needs to be done to implement in each language, we measure the number of characters of non-whitespace, non-comment code in each implementation.
All implementations were written in natural Python or C++, and a conscious attempt was made to follow the DyNet Python implementation where possible.
\item[Accuracy:]
In order to confirm that the code in each library is doing the same thing, we compared accuracies and ensure that they are in the same general range with only small differences, and thus do not include these statistics in our main comparison.
However, in some cases we will show accuracies for each task: per-word negative log likelihood for the RNNLM, tagging accuracy for the BiLSTM tagger, and fine-grained sentiment classification accuracy at the root for the Tree LSTM.
\end{description}

\subsection{Evaluation Results}
\label{sec:evalresults}

\subsubsection{Cross-toolkit Comparison}
\label{sec:expcomparison}

First, we perform a comparison of computation speed over the four tasks across the DyNet Python interface, the DyNet C++ interface, Chainer, Theano, and TensorFlow.
Additionally, as discussed in \secref{rnnbuilder}, DyNet also has experimental support for efficient sequence-level processing such as that implemented in Theano and TensorFlow, so we present numbers for this on the RNNLM task as well (labeled DyC++ Seq) to examine the effect of sharing computations at the sequence level.
We also vary the mini-batch size of the RNNLM to demonstrate the effect of mini-batching.
The results are shown in \tabref{speedscpu} and \tabref{speedsgpu}.

\begin{table}
\begin{center}
\scalebox{0.9}{
\begin{tabular}{l|c|rrr|rrr}
 & Metric & DyC++ & DyPy & Chainer & DyC++ Seq & Theano & TF \\ \hline
RNNLM (MB=1) & words/sec & 190 & 190 & 114 & 494 & 189 & 298 \\
RNNLM (MB=4) & words/sec & 830 & 825 & 295 & 1510 & 567 & 473 \\
RNNLM (MB=16) & words/sec & 1820 & 1880 & 794 & 2400 & 1100 & 606 \\
RNNLM (MB=64) & words/sec & 2440 & 2470 & 1340 & 2820 & 1260 & 636 \\
\hline
BiLSTM Tag \phantom{-sparse} & words/sec & 427 & 428 & 22.7 & \multicolumn{1}{c}{-} & 102 & 143 \\
BiLSTM Tag +sparse & words/sec & 8410 & 7990 & \multicolumn{1}{c|}{-} & \multicolumn{1}{c}{-} & \multicolumn{1}{c}{-} & \multicolumn{1}{c}{-} \\

BiLSTM Tag+Char \phantom{-sparse} & words/sec & 419 & 413 & 22.0 & \multicolumn{1}{c}{-} & 94.3 & \multicolumn{1}{c}{-} \\
BiLSTM Tag+Char +sparse & words/sec & 6530 & 6320 & \multicolumn{1}{c|}{-} & \multicolumn{1}{c}{-} & \multicolumn{1}{c}{-} & \multicolumn{1}{c}{-} \\

\hline
TreeLSTM \phantom{-sparse} & sents/sec & 91.6 & 88.1 & 7.21 & \multicolumn{1}{c}{-} & \multicolumn{1}{c}{-} & \multicolumn{1}{c}{-} \\
TreeLSTM +sparse & sents/sec & 186 & 173 & \multicolumn{1}{c|}{-} & \multicolumn{1}{c}{-} & \multicolumn{1}{c}{-} & \multicolumn{1}{c}{-} \\

\end{tabular}}
\end{center}
\caption{Processing speed for each toolkit on \textbf{CPU}. Speeds are measured in words/sec for RNNLM and Tagger and sentences/sec for TreeLSTM. Lines with +sparse indicate sparse updates for the LookupParameters, which is the default behavior in DyNet, but not comparable to the implementations in other toolkits, which are performing dense updates. }
\label{tab:speedscpu}
\end{table}

\textbf{Comparison on CPU:}
First focusing on the CPU results, we can see that DyNet handily outperforms the other toolkits in efficiency.
This is true for the more standard and straightforward RNNLMs (where speeds are 1.66x to 3.20x faster than the fastest counterpart), particularly for the more complicated tasks such as the BiLSTM tagger (gains of 2.99x to 4.44x) and TreeLSTM (a gain of 12.7x).
This is a result of the DyNet design described in \secref{design}, which focuses on minimizing overhead in graph construction and focuses on optimizing for speed on both CPU and GPU.

In addition, comparing the various DyNet interfaces, we see that there is a negligible difference between the C++ and Python interfaces, due to the fact that Python simply places a thin wrapper of the core C++ code.
On the other hand, utilizing the sequence-based computation interface provides significant improvements, particularly at smaller RNNLM batch sizes.
This demonstrates the utility of sharing computations across different time steps, particularly when fewer computations are shared across training instances.%
\footnote{Recently, \cite{looks2017dynamic} have proposed a method for dynamically combining similar calculations in computation graphs, and it is possible that by applying this or similar methods to computation graphs generated by the canonical DyNet implementation, this gap could be closed.}

Next, we turn to the +sparse results, which use the sparse updates that are described in \secref{sparseupdates} and are on by default in DyNet.
From these results, we can see that we obtain further large gains in speed, particularly for the BiLSTM tagger, which demonstrated gains of up to 19.7x over dense updates.
While these results are not directly comparable to the dense updates used in the implementations in the other toolkits, they do demonstrate that further large speed gains can be achieved through the use of the sparse update strategy.
We present more results and analysis comparing dense and sparse update strategies on both CPU and GPU in \secref{expsparseupdates}.

\begin{table}
\begin{center}
\scalebox{0.9}{
\begin{tabular}{c|c|rrr|rrr}
 &Metric & DyC++ & DyPy & Chainer & DyC++ Seq & Theano & TF \\ \hline
RNNLM (MB=1) & words/sec & 1060 & 1060 & 290 & 1810 & 1040 & 1030 \\
RNNLM (MB=4) & words/sec & 1660 & 1660 & 972 & 3980 & 2560 & 3830 \\
RNNLM (MB=16) & words/sec & 5730 & 5740 & 3810 & 10400 & 6370 & 13900 \\
RNNLM (MB=64) & words/sec & 18500 & 18500 & 14400 & 21200 & 6420 & 37000 \\
\hline
BiLSTM Tag & words/sec & 1250 & 1250 & 147 & \multicolumn{1}{c}{-} & 823 & 492 \\
BiLSTM Tag+Char & words/sec & 1010 & 1010 & 122 & \multicolumn{1}{c}{-} & 434 & \multicolumn{1}{c}{-} \\
\hline
TreeLSTM & sents/sec & 73.3 & 73.5 & 9.27 & \multicolumn{1}{c}{-} & \multicolumn{1}{c}{-} & \multicolumn{1}{c}{-} \\
\end{tabular}}
\end{center}
\caption{Processing speed for each toolkit on \textbf{GPU}. Speeds are measured in words/sec for RNNLM and Tagger and sentences/sec for TreeLSTM.}
\label{tab:speedsgpu}
\end{table}

\textbf{Comparison on GPU:}
Next, we discuss the results when performing computation on GPU.
From the figure, we can see that for RNNLM with smaller batch sizes, and for all settings of the BiLSTM Tagger and TreeLSTM tasks, DyNet significantly outperforms the other implementations, although the gap is smaller than on CPU.
On the other hand, for large batch sizes of the RNNLM, TensorFlow scales well in the size of the minibatch, outperforming DyNet after a mini-batch size of 16, likely a result of its highly optimized methods to efficiently schedule computation on the GPU.
Regardless, the results for DyNet are still competitive, outperforming the other baseline toolkits, and given the other advantages of the dynamic declaration paradigm (intuitive programming and debugging), we believe it presents an attractive alternative even in situations where static declaration toolkits are highly optimized.

It is also interesting to compare the speed results for CPU and GPU.
It is notable that for the relatively simple RNN with large mini-batches, computation on the GPU is much more efficient, as mini-batches get smaller or models get more complicated, this advantage shrinks, and for tree-based networks, the implementation on CPU is more efficient.
This stems from the fact that GPUs excel at performing large operations, but for smaller nets the overhead for launching operations on the GPU is large enough that computation on the GPU is not beneficial.
In addition, as we will show in \secref{expsparseupdates}, sparse updates provide limited benefit to GPU processing, and thus for the BiLSTM and TreeLSTM, sparse updates on the CPU are the most efficient training strategy.

\textbf{Comparison of Startup Time:}
Finally, we compare startup time between the toolkits, as shown in \tabref{startup}.
From the results we can see that in general the dynamic declaration toolkits DyNet and Chainer tend to have shorter startup time, as there is no separate graph compilation step.
In particular, the overhead for this compilation step is large, and increases as models get more complicated to taking over a minute and a half for the BiLSTM tagger with character features.\footnote{This does not include an additional step necessary to compile operations used in the graph, which required approximately an additional 2 minutes, but can be cached on succeeding runs.}

\begin{table}
\begin{center}
\begin{tabular}{c|rrr|rr}
 & DyC++ & DyPy & Chainer & Theano & TF \\ \hline
RNNLM & 1.29 & 1.41 & 0.727 & 17.2 & 2.73 \\
BiLSTM Tag & 5.64 & 8.45 & 7.23 & 46.4 & 9.16 \\
BiLSTM Tag+Char & 7.18 & 9.90 & 9.03 & 93.1 & \multicolumn{1}{c}{-} \\
TreeLSTM & 1.94 & 5.42 & 4.86 & \multicolumn{1}{c}{-} & \multicolumn{1}{c}{-} \\
\end{tabular}
\end{center}
\caption{Time from program start to processing the first instance for each toolkit (seconds).}
\label{tab:startup}
\end{table}

\subsubsection{Effect of Sparse Updates}
\label{sec:expsparseupdates}

\begin{table}
\begin{center}
\begin{tabular}{c|rr|rr|rr|rr}
 & \multicolumn{4}{c|}{Speed} & \multicolumn{4}{c}{Accuracy} \\
 & \multicolumn{2}{c|}{Dense} & \multicolumn{2}{c|}{Sparse} & \multicolumn{2}{c|}{Dense} & \multicolumn{2}{c}{Sparse} \\
 & CPU & GPU & CPU & GPU & CPU & GPU & CPU & GPU \\ \hline
RNNLM (MB=1)  & 190 & 1060 & 205 & 1070 & -5.55 & -5.22 & -5.65 & -5.29 \\
RNNLM (MB=16) & 1820 & 5730 & 1920 & 5320 & -5.19 & -4.86 & -5.18 & -4.89 \\
BiLSTM Tag & 427 & 1250 & 8410 & 1700 & 92.4 & 93.2 & 94.1 & 92.9 \\
BiLSTM Tag+Char & 419 & 1010 & 6530 & 1290 & 93.4 & 93.9 & 94.6 & 93.4 \\
TreeLSTM & 91.6 & 73.3 & 186 & 76.3 & 47.2 & 47.6 & 48.1 & 47.1 \\
\end{tabular}
\end{center}
\caption{Processing speed and accuracy after 10 minutes with dense or sparse updates.}
\label{tab:sparseresults}
\end{table}

Next, we discuss in detail the effect of sparse updates described in \secref{sparseupdates}, with results shown in \tabref{sparseresults}.
First concentrating on the CPU results, we can see that we can achieve uniform speed improvements, although the improvement greatly varies, from a mere 1.05x speedup for RNNLM with minibatch size of 16, to a 20x improvement for the BiLSTM tagger.
The difference between tasks can be attributed to the sparseness of the word embeddings and their size compared to the other parts of the computation graph.
In the RNNLM, in addition to the input word embeddings, there is also a large softmax layer that is updated in a dense fashion at every time step, so even with sparse updates on the input site, there is a significant amount of other calculation that prevents large speedups.\footnote{Results would likely be different if using a more sufficient softmax implementation such as those described in \secref{softmaxbuilder}}
On the other hand, the BiLSTM tagger has a large input vocabulary, but only a small output vocabulary of 9 tags, and thus preventing excess calculation over the input is effective.
Looking at accuracy after training for 10 minutes, it is clear that CPU with sparse updates gives the best accuracy of the various settings for the BiLSTM tagger and TreeLSTM, indicating that this may be a good choice for similar natural language analysis tasks.

On the other hand, on GPU the training speeds are only nominally faster for sparse updates in most cases, and even slower in some cases.
From these results, we can conclude that sparse updates are an attractive option for tasks in which there are large embedding matrices and CPU training is an option, but less so for training on GPU.

\subsubsection{Code Complexity}
\label{sec:expcomplexity}

\begin{table}
\begin{center}
\begin{tabular}{c|rrr|rrr}
 & DyC++ & DyPy & Chainer & DyC++ Seq & Theano & TF \\ \hline
RNNLM & 4744 & 3766 & 3553 & 4659 & 4257 & 4922 \\
BiLSTM Tag & 5245 & 4532 & 4412 & \multicolumn{1}{c}{-} & 5581 & 5137 \\
BiLSTM Tag+Char & 6180 & 5324 & 5028 & \multicolumn{1}{c}{-} & 5636 & \multicolumn{1}{c}{-} \\
TreeLSTM & 6869 & 5242 & 5492 & \multicolumn{1}{c}{-} & \multicolumn{1}{c}{-} & \multicolumn{1}{c}{-} \\
\end{tabular}
\end{center}
\caption{Number of non-comment characters in the implementation of each toolkit.}
\label{tab:complexity}
\end{table}

Finally, to give a very rough estimate of the amount of effort required to implement tasks in each toolkit, we compare the number of non-comment characters in each implementation, with the results shown in \tabref{complexity}.
It should be noted that all of the implementations are written to be natural and readable, with no effort to reduce the number of characters therein.

From the results we can see that among the programs implemented in python (DyPy, Chainer, Theano, TF), both the DyNet and Chainer implementations are consistently shorter than the Theano and TensorFlow implementations, an indication that the dynamic declaration paradigm allows for simpler implementation in fewer characters of code.
This conciseness is orthogonal the previously mentioned advantages of dynamically declared graphs being able to write programs in a form that is more intuitive and closer to the standard API, which is not directly reflected in these numbers.

%% file: 055-usecases.tex
DyNet is already in active use, and has been used for a wide variety of projects, mostly related to natural language processing.
DyNet itself contains a number of examples (in the \texttt{examples/} directory) of minimal to moderate complexity.
We also list a number of full-scale research projects, to allow interested readers to find reference implementations that match their application of interest.

\begin{description}
\item[Syntactic Parsing:] Parsing is currently the most prominent scenario in which DyNet has been used, and DyNet was behind the development of a number of methods such as stack LSTMs \cite{dyer2015stacklstm} (\url{https://github.com/clab/lstm-parser}), bi-directional LSTM feature extractors for dependency parsing \cite{kiperwasser2016bilstmparser} (\url{https://github.com/elikip/bist-parser}), recurrent neural network grammars \cite{dyer2016rnng} (\url{https://github.com/clab/rnng}), and hierarchical tree LSTMs \cite{kiperwasser2016eftreelstm} (\url{https://github.com/elikip/htparser}).
\item[Machine Translation:] DyNet has contributed to creation of methods for incorporating biases in attention \cite{cohn2016alignment} (\url{https://github.com/trevorcohn/mantis}) character-based methods for translation \cite{ling2015character}. It also powers a number of machine translation toolkits such as Lamtram (\url{https://github.com/neubig/lamtram}) and nmtkit (\url{https://github.com/odashi/nmtkit}).
\item[Language Modeling:] DyNet has been used in the development of hybrid neural/$n$-gram language models \cite{neubig2016modlm} (\url{https://github.com/neubig/modlm}), and generative syntactic language models \cite{dyer2016rnng} (\url{https://github.com/clab/rnng}).
\item[Tagging:] DyNet was used in development of methods for named entity recognition \cite{lample2016neural} (\url{https://github.com/clab/stack-lstm-ner}), POS-tagging \cite{plank16tagging}, semantic role labeling \cite{swayamdipta2016semantic} (\url{https://github.com/clab/joint-lstm-parser}), punctuation prediction \cite{ballesteros2016punctuation} (\url{https://github.com/miguelballesteros/LSTM-punctuation}), and multi-task learning for sequence processing \cite{klerke16gaze,sogaard16mtl}, as well as creation of new architectures such as segmental recurrent neural networks \cite{kong2016segrnn}(\url{https://github.com/clab/dynet/tree/master/examples/cpp/segrnn-sup}).
\item[Morphology:] DyNet has been used for morphological inflection generation \cite{faruqui2016morphology,aharoni16morph} \\ (\url{https://github.com/mfaruqui/morph-trans} \\ \url{https://github.com/roeeaharoni/morphological-reinflection}).
\item[Misc:] DyNet has been used developing specialized networks for detection of coordination structures \cite{ficler16coord}; semi-supervised preposition-sense disambiguation \cite{gonen16semisup}; and for identifying lexical semantic relations \cite{shwartz16hypenet,shwartz16lexnet} (\url{https://github.com/vered1986/HypeNET}).
\end{description}

%% file: 060-conclusion.tex
In this paper, we introduced DyNet, a toolkit designed for creation of dynamic neural networks such as those commonly used in natural language processing.
As shown in the experimental evaluation, DyNet has a number of advantages over existing toolkits in conceptual simplicity and ability to handle more complicated neural network structures with a minimum of overhead.

The work is far from complete, however, and we have a number of projects in store for the future:
\begin{description}
\item[Multi-device Support:]
Currently, DyNet supports execution on a single GPU or CPU, but does not allow model parallelism.
We plan to introduce the ability to execute a single computation graph on multiple devices in the near future.
\item[On-the-fly Graph Optimization:]
As mentioned in \secref{expcomparison}, DyNet's optimized sequence-based interface shows improvements because it is able to combine together operations of similar shapes.
An interesting challenge is functionality to do this automatically, on the fly, which would allow users to program in the intuitive canonical DyNet interface but still reap the benefit of combining similar operations.
\item[Support for Operations and Optimizers:]
DyNet continues to grow its library of supported operations and optimizers to allow implementation of new methods proposed in the literature.
\end{description}
Finally, DyNet is a community effort, and we welcome creative contributions from the community to make it a better tool for all.

%% file: 070-acks.tex
DyNet has been developed by a community of users, and as such, it has been supported indirectly by numerous funding sources. The initial development of DyNet (when it was called cnn) was supported in large part by the U. S. Army Research Laboratory and the U. S. Army Research Office under contract/grant number W911NF-10-1-0533, and continued development has been supported in large part by JSPS KAKENHI Grant Number 16H05873. The Python bindings development was supported by the Israeli Science Foundation (grant number 1555/15).

%% file: main.bbl
\begin{thebibliography}{10}

\bibitem{abadi2016tensorflow}
Mart{\i}n Abadi, Ashish Agarwal, Paul Barham, Eugene Brevdo, Zhifeng Chen,
  Craig Citro, Greg~S Corrado, Andy Davis, Jeffrey Dean, Matthieu Devin, et~al.
\newblock Tensorflow: Large-scale machine learning on heterogeneous distributed
  systems.
\newblock {\em arXiv preprint arXiv:1603.04467}, 2016.

\bibitem{aharoni16morph}
Roee Aharoni, Yoav Goldberg, and Yonatan Belinkov.
\newblock Improving sequence to sequence learning for morphological inflection
  generation: The {BIU-MIT} systems for the {SIGMORPHON} 2016 shared task for
  morphological reinflection.
\newblock pages 41--48, 2016.

\bibitem{andreas2016compose}
Jacob Andreas, Marcus Rohrbach, Trevor Darrell, and Dan Klein.
\newblock Learning to compose neural networks for question answering.
\newblock In {\em Conference of the North American Chapter of the Association
  for Computational Linguistics (NAACL)}, pages 1545--1554, 2016.

\bibitem{bahdanau15alignandtranslate}
Dzmitry Bahdanau, Kyunghyun Cho, and Yoshua Bengio.
\newblock Neural machine translation by jointly learning to align and
  translate.
\newblock In {\em International Conference on Learning Representations (ICLR)},
  2015.

\bibitem{ballesteros2016punctuation}
Miguel Ballesteros and Leo Wanner.
\newblock A neural network architecture for multilingual punctuation
  generation.
\newblock In {\em Conference on Empirical Methods in Natural Language
  Processing (EMNLP)}, pages 1048--1053, 2016.

\bibitem{bengio2003neural}
Yoshua Bengio, R{\'e}jean Ducharme, Pascal Vincent, and Christian Jauvin.
\newblock A neural probabilistic language model.
\newblock {\em Journal of Machine Learning Research}, 3(Feb):1137--1155, 2003.

\bibitem{bergstra2010theano}
James Bergstra, Olivier Breuleux, Fr{\'e}d{\'e}ric Bastien, Pascal Lamblin,
  Razvan Pascanu, Guillaume Desjardins, Joseph Turian, David Warde-Farley, and
  Yoshua Bengio.
\newblock Theano: A {CPU} and {GPU} math compiler in {Python}.
\newblock In {\em Proc. 9th Python in Science Conf}, pages 1--7, 2010.

\bibitem{bowman2016spinn}
Samuel~R. Bowman, Jon Gauthier, Abhinav Rastogi, Raghav Gupta, Christopher~D.
  Manning, and Christopher Potts.
\newblock A fast unified model for parsing and sentence understanding.
\newblock In {\em Annual Conference of the Association for Computational
  Linguistics (ACL)}, pages 1466--1477, 2016.

\bibitem{buckman2016backtracking}
Jacob Buckman, Miguel Ballesteros, and Chris Dyer.
\newblock Transition-based dependency parsing with heuristic backtracking.
\newblock In {\em Conference on Empirical Methods in Natural Language
  Processing (EMNLP)}, pages 2313--2318, 2016.

\bibitem{chen2015mxnet}
Tianqi Chen, Mu~Li, Yutian Li, Min Lin, Naiyan Wang, Minjie Wang, Tianjun Xiao,
  Bing Xu, Chiyuan Zhang, and Zheng Zhang.
\newblock Mxnet: A flexible and efficient machine learning library for
  heterogeneous distributed systems.
\newblock {\em arXiv preprint arXiv:1512.01274}, 2015.

\bibitem{chung2014gru}
Junyoung Chung, Caglar Gulcehre, KyungHyun Cho, and Yoshua Bengio.
\newblock Empirical evaluation of gated recurrent neural networks on sequence
  modeling.
\newblock {\em arXiv preprint arXiv:1412.3555}, 2014.

\bibitem{cohn2016alignment}
Trevor Cohn, Cong Duy~Vu Hoang, Ekaterina Vymolova, Kaisheng Yao, Chris Dyer,
  and Gholamreza Haffari.
\newblock Incorporating structural alignment biases into an attentional neural
  translation model.
\newblock In {\em Conference of the North American Chapter of the Association
  for Computational Linguistics (NAACL)}, pages 876--885, 2016.

\bibitem{collobert2002torch}
Ronan Collobert, Samy Bengio, and Johnny Mari{\'e}thoz.
\newblock Torch: a modular machine learning software library.
\newblock Technical report, Idiap, 2002.

\bibitem{collobert2011natural}
Ronan Collobert, Jason Weston, L{\'e}on Bottou, Michael Karlen, Koray
  Kavukcuoglu, and Pavel Kuksa.
\newblock Natural language processing (almost) from scratch.
\newblock {\em Journal of Machine Learning Research}, 12(Aug):2493--2537, 2011.

\bibitem{do:2010}
Trinh-Minh-Tri Do and Thierry Arti\`{e}res.
\newblock Neural conditional random fields.
\newblock In {\em Proceedings of the Thirteenth International Conference on
  Artificial Intelligence and Statistics}, 2010.

\bibitem{duchi2011adaptive}
John Duchi, Elad Hazan, and Yoram Singer.
\newblock Adaptive subgradient methods for online learning and stochastic
  optimization.
\newblock {\em Journal of Machine Learning Research}, 12(Jul):2121--2159, 2011.

\bibitem{dyer2015stacklstm}
Chris Dyer, Miguel Ballesteros, Wang Ling, Austin Matthews, and Noah~A. Smith.
\newblock Transition-based dependency parsing with stack long short-term
  memory.
\newblock In {\em Annual Conference of the Association for Computational
  Linguistics (ACL)}, pages 334--343, 2015.

\bibitem{dyer2016rnng}
Chris Dyer, Adhiguna Kuncoro, Miguel Ballesteros, and Noah~A. Smith.
\newblock Recurrent neural network grammars.
\newblock In {\em Conference of the North American Chapter of the Association
  for Computational Linguistics (NAACL)}, pages 199--209, 2016.

\bibitem{eisner:2005}
Jason Eisner, Eric Goldlust, and Noah~A. Smith.
\newblock Compiling comp ling: Practical weighted dynamic programming and the
  {Dyna} language.
\newblock In {\em Conference on Empirical Methods in Natural Language
  Processing (EMNLP)}, 2005.

\bibitem{elman90rnn}
Jeffrey~L Elman.
\newblock Finding structure in time.
\newblock {\em Cognitive science}, 14(2):179--211, 1990.

\bibitem{faruqui2016morphology}
Manaal Faruqui, Yulia Tsvetkov, Graham Neubig, and Chris Dyer.
\newblock Morphological inflection generation using character sequence to
  sequence learning.
\newblock In {\em Conference of the North American Chapter of the Association
  for Computational Linguistics (NAACL)}, pages 634--643, 2016.

\bibitem{ficler16coord}
Jessica Ficler and Yoav Goldberg.
\newblock A neural network for coordination boundary prediction.
\newblock In {\em Conference on Empirical Methods in Natural Language
  Processing (EMNLP)}, pages 23--32, 2016.

\bibitem{gonen16semisup}
Hila Gonen and Yoav Goldberg.
\newblock Semi supervised preposition-sense disambiguation using multilingual
  data.
\newblock In {\em International Conference on Computational Linguistics
  (COLING)}, pages 2718--2729, 2016.

\bibitem{goodman2001classes}
Joshua Goodman.
\newblock Classes for fast maximum entropy training.
\newblock In {\em IEEE International Conference on Acoustics, Speech, and
  Signal Processing (ICASSP)}, volume~1, pages 561--564. IEEE, 2001.

\bibitem{gormley:2015}
Matthew~R. Gormley, Mark Dredze, and Jason Eisner.
\newblock Approximation-aware dependency parsing by belief propagation.
\newblock {\em Transactions of the Association for Computational Linguistics},
  3, 2015.

\bibitem{graves:2006}
Alex Graves, Santiago Fern\'{a}ndez, Faustino Gomez, and J\"{u}rgen
  Schmidhuber.
\newblock Connectionist temporal classification: Labelling unsegmented sequence
  data with recurrent neural networks.
\newblock In {\em International Conference on Machine Learning (ICML)}, 2006.

\bibitem{griewank1991automatic}
Andreas Griewank.
\newblock Automatic differentiation of algorithms: theory, implementation, and
  application.
\newblock In {\em proceedings of the first SIAM Workshop on Automatic
  Differentiation}, 1991.

\bibitem{guennebaud2010eigen}
Ga\"{e}l Guennebaud, Beno\^{i}t Jacob, et~al.
\newblock Eigen v3.
\newblock http://eigen.tuxfamily.org, 2010.

\bibitem{hinton2012deep}
Geoffrey Hinton, Li~Deng, Dong Yu, George~E Dahl, Abdel-rahman Mohamed, Navdeep
  Jaitly, Andrew Senior, Vincent Vanhoucke, Patrick Nguyen, Tara~N Sainath,
  et~al.
\newblock Deep neural networks for acoustic modeling in speech recognition: The
  shared views of four research groups.
\newblock {\em IEEE Signal Processing Magazine}, 29(6):82--97, 2012.

\bibitem{hochreiter97lstm}
Sepp Hochreiter and J{\"u}rgen Schmidhuber.
\newblock Long short-term memory.
\newblock {\em Neural computation}, 9(8):1735--1780, 1997.

\bibitem{hogan2014fast}
Robin~J. Hogan.
\newblock Fast reverse-mode automatic differentiation using expression
  templates in {C++}.
\newblock {\em ACM Transactions on Mathematical Software (TOMS)}, 40(4):26,
  2014.

\bibitem{huang2015bidirectional}
Zhiheng Huang, Wei Xu, and Kai Yu.
\newblock Bidirectional {LSTM-CRF} models for sequence tagging.
\newblock {\em arXiv preprint arXiv:1508.01991}, 2015.

\bibitem{iyyer2014factoid}
Mohit Iyyer, Jordan Boyd-Graber, Leonardo Claudino, Richard Socher, and Hal
  Daum\'{e}~III.
\newblock A neural network for factoid question answering over paragraphs.
\newblock In {\em Conference on Empirical Methods in Natural Language
  Processing (EMNLP)}, pages 633--644, 2014.

\bibitem{kingma2014adam}
Diederik Kingma and Jimmy Ba.
\newblock Adam: A method for stochastic optimization.
\newblock {\em arXiv preprint arXiv:1412.6980}, 2014.

\bibitem{kiperwasser2016eftreelstm}
Eliyahu Kiperwasser and Yoav Goldberg.
\newblock Easy-first dependency parsing with hierarchical tree {LSTMs}.
\newblock {\em Transactions of the Association for Computational Linguistics},
  4:445--461, 2016.

\bibitem{kiperwasser2016bilstmparser}
Eliyahu Kiperwasser and Yoav Goldberg.
\newblock Simple and accurate dependency parsing using bidirectional {LSTM}
  feature representations.
\newblock {\em Transactions of the Association for Computational Linguistics},
  4:313--327, 2016.

\bibitem{klerke16gaze}
Sigrid Klerke, Yoav Goldberg, and Anders S{\o}gaard.
\newblock Improving sentence compression by learning to predict gaze.
\newblock In {\em Conference of the North American Chapter of the Association
  for Computational Linguistics (NAACL)}, pages 1528--1533, 2016.

\bibitem{kong2016segrnn}
Lingpeng Kong, Chris Dyer, and Noah~A. Smith.
\newblock Segmental recurrent neural networks.
\newblock In {\em International Conference on Learning Representations (ICLR)},
  2016.

\bibitem{krizhevsky2012imagenet}
Alex Krizhevsky, Ilya Sutskever, and Geoffrey~E Hinton.
\newblock Imagenet classification with deep convolutional neural networks.
\newblock In {\em Neural Information Processing Systems (NIPS)}, pages
  1097--1105, 2012.

\bibitem{lample2016neural}
Guillaume Lample, Miguel Ballesteros, Sandeep Subramanian, Kazuya Kawakami, and
  Chris Dyer.
\newblock Neural architectures for named entity recognition.
\newblock In {\em Conference of the North American Chapter of the Association
  for Computational Linguistics (NAACL)}, pages 260--270, 2016.

\bibitem{liang:2016eccv}
Xiaodan Liang, Xiaohui Shen, Jiashi Feng, Liang Lin, and Shuicheng Yan.
\newblock Semantic object parsing with graph lstm.
\newblock 2016.

\bibitem{ling2015character}
Wang Ling, Isabel Trancoso, Chris Dyer, and Alan~W Black.
\newblock Character-based neural machine translation.
\newblock {\em arXiv preprint arXiv:1511.04586}, 2015.

\bibitem{looks2017dynamic}
Moshe Looks, Marcello Herreshoff, DeLesley Hutchins, and Peter Norvig.
\newblock Deep learning with dynamic computation graphs.
\newblock In {\em Submitted to International Conference on Learning
  Representations (ICLR)}, 2017.

\bibitem{mikolov2011extensions}
Tom{\'a}{\v{s}} Mikolov, Stefan Kombrink, Luk{\'a}{\v{s}} Burget, Jan
  {\v{C}}ernock{\`y}, and Sanjeev Khudanpur.
\newblock Extensions of recurrent neural network language model.
\newblock In {\em IEEE International Conference on Acoustics, Speech, and
  Signal Processing (ICASSP)}, pages 5528--5531. IEEE, 2011.

\bibitem{mnih2015human}
Volodymyr Mnih, Koray Kavukcuoglu, David Silver, Andrei~A Rusu, Joel Veness,
  Marc~G Bellemare, Alex Graves, Martin Riedmiller, Andreas~K Fidjeland, Georg
  Ostrovski, et~al.
\newblock Human-level control through deep reinforcement learning.
\newblock {\em Nature}, 518(7540):529--533, 2015.

\bibitem{nesterov1983method}
Yurii Nesterov.
\newblock A method of solving a convex programming problem with convergence
  rate o (1/k2).
\newblock In {\em Soviet Mathematics Doklady}, volume~27, pages 372--376, 1983.

\bibitem{neubig2016modlm}
Graham Neubig and Chris Dyer.
\newblock Generalizing and hybridizing count-based and neural language models.
\newblock In {\em Conference on Empirical Methods in Natural Language
  Processing (EMNLP)}, pages 1163--1172, 2016.

\bibitem{nothman2012wikiner}
Joel Nothman, Nicky Ringland, Will Radford, Tara Murphy, and James~R. Curran.
\newblock Learning multilingual named entity recognition from {Wikipedia}.
\newblock {\em Artificial Intelligence}, 194:151--175, 2012.

\bibitem{plank16tagging}
Barbara Plank, Anders S{\o}gaard, and Yoav Goldberg.
\newblock Multilingual part-of-speech tagging with bidirectional long
  short-term memory models and auxiliary loss.
\newblock In {\em Annual Conference of the Association for Computational
  Linguistics (ACL)}, pages 412--418, 2016.

\bibitem{recht2011hogwild}
Benjamin Recht, Christopher Re, Stephen Wright, and Feng Niu.
\newblock Hogwild: A lock-free approach to parallelizing stochastic gradient
  descent.
\newblock In {\em Neural Information Processing Systems (NIPS)}, pages
  693--701, 2011.

\bibitem{shalev2011pegasos}
Shai Shalev-Shwartz, Yoram Singer, Nathan Srebro, and Andrew Cotter.
\newblock Pegasos: Primal estimated sub-gradient solver for {SVM}.
\newblock {\em Mathematical programming}, 127(1):3--30, 2011.

\bibitem{shwartz16lexnet}
Vered Shwartz and Ido Dagan.
\newblock Cogalex-v shared task: Lexnet - integrated path-based and
  distributional method for the identification of semantic relations.
\newblock In {\em Proceedings of the 5th Workshop on Cognitive Aspects of the
  Lexicon (CogALex - V)}, pages 80--85, 2016.

\bibitem{shwartz16hypenet}
Vered Shwartz, Yoav Goldberg, and Ido Dagan.
\newblock Improving hypernymy detection with an integrated path-based and
  distributional method.
\newblock In {\em Annual Conference of the Association for Computational
  Linguistics (ACL)}, pages 2389--2398, 2016.

\bibitem{silver2016mastering}
David Silver, Aja Huang, Chris~J Maddison, Arthur Guez, Laurent Sifre, George
  Van Den~Driessche, Julian Schrittwieser, Ioannis Antonoglou, Veda
  Panneershelvam, Marc Lanctot, et~al.
\newblock Mastering the game of {Go} with deep neural networks and tree search.
\newblock {\em Nature}, 529(7587):484--489, 2016.

\bibitem{socher11recursivenn}
Richard Socher, Cliff~C Lin, Chris Manning, and Andrew~Y Ng.
\newblock Parsing natural scenes and natural language with recursive neural
  networks.
\newblock In {\em International Conference on Machine Learning (ICML)}, pages
  129--136, 2011.

\bibitem{sogaard16mtl}
Anders S{\o}gaard and Yoav Goldberg.
\newblock Deep multi-task learning with low level tasks supervised at lower
  layers.
\newblock In {\em Proceedings of the 54th Annual Meeting of the Association for
  Computational Linguistics (Volume 2: Short Papers)}, pages 231--235, 2016.

\bibitem{sun2016asynchronous}
Xu~Sun.
\newblock Asynchronous parallel learning for neural networks and structured
  models with dense features.
\newblock In {\em International Conference on Computational Linguistics
  (COLING)}, pages 192--202, 2016.

\bibitem{sundermeyer12lstmlm}
Martin Sundermeyer, Ralf Schl{\"u}ter, and Hermann Ney.
\newblock {LSTM} neural networks for language modeling.
\newblock In {\em Annual Conference of the International Speech Communication
  Association (InterSpeech)}, 2012.

\bibitem{sutskever14sequencetosequence}
Ilya Sutskever, Oriol Vinyals, and Quoc~VV Le.
\newblock Sequence to sequence learning with neural networks.
\newblock In {\em Neural Information Processing Systems (NIPS)}, pages
  3104--3112, 2014.

\bibitem{swayamdipta2016semantic}
Swabha Swayamdipta, Miguel Ballesteros, Chris Dyer, and Noah~A. Smith.
\newblock Greedy, joint syntactic-semantic parsing with stack lstms.
\newblock In {\em Conference on Natural Language Learning (CoNLL)}, pages
  187--197, 2016.

\bibitem{tai15treelstm}
Kai~Sheng Tai, Richard Socher, and Christopher~D. Manning.
\newblock Improved semantic representations from tree-structured long
  short-term memory networks.
\newblock In {\em Annual Conference of the Association for Computational
  Linguistics (ACL)}, 2015.

\bibitem{tokui2015chainer}
Seiya Tokui, Kenta Oono, Shohei Hido, and Justin Clayton.
\newblock Chainer: a next-generation open source framework for deep learning.
\newblock In {\em Proceedings of Workshop on Machine Learning Systems
  (LearningSys) in The Twenty-ninth Annual Conference on Neural Information
  Processing Systems (NIPS)}, 2015.

\bibitem{wengert1964asa}
R.E. Wengert.
\newblock A simple automatic derivative evaluation program.
\newblock {\em Communications of the {ACM}}, 7(8):463--464, 1964.

\bibitem{yu2014cntk}
Dong Yu, Adam Eversole, Mike Seltzer, Kaisheng Yao, Oleksii Kuchaiev, Yu~Zhang,
  Frank Seide, Zhiheng Huang, Brian Guenter, Huaming Wang, Jasha Droppo,
  Geoffrey Zweig, Chris Rossbach, Jie Gao, Andreas Stolcke, Jon Currey, Malcolm
  Slaney, Guoguo Chen, Amit Agarwal, Chris Basoglu, Marko Padmilac, Alexey
  Kamenev, Vladimir Ivanov, Scott Cypher, Hari Parthasarathi, Bhaskar Mitra,
  Baolin Peng, and Xuedong Huang.
\newblock An introduction to computational networks and the computational
  network toolkit.
\newblock Technical report, Microsoft Research, October 2014.

\bibitem{zen2013statistical}
Heiga Zen, Andrew Senior, and Mike Schuster.
\newblock Statistical parametric speech synthesis using deep neural networks.
\newblock In {\em IEEE International Conference on Acoustics, Speech, and
  Signal Processing (ICASSP)}, pages 7962--7966. IEEE, 2013.

\end{thebibliography}
